\documentclass[manuscript,screen]{acmart}
\AtBeginDocument{%
  }

\setcopyright{acmlicensed}
\copyrightyear{2026}
\acmYear{2026}
\acmDOI{XXXXXXX.XXXXXXX}
\acmJournal{TOSEM}

\usepackage{xcolor}
\usepackage{amsmath}
\usepackage{tabularx}
\usepackage{multirow}
\usepackage{tcolorbox}
\usepackage{graphicx}
\usepackage{subcaption}
\usepackage{enumitem}
\usepackage{colortbl}
\usepackage{pifont}
\usepackage{balance} 
\usepackage{flushend} 
\usepackage{adjustbox}
\usepackage{booktabs,array}
\tcbuselibrary{breakable}
\usepackage{listings}
\usepackage{tikz}
\tcbuselibrary{skins,breakable}

\definecolor{ragq}{HTML}{185FA5}   
\definecolor{ragc}{HTML}{0F6E56}   
\definecolor{ragr}{HTML}{993C1D}   
\definecolor{ragx}{HTML}{534AB7}   
\definecolor{ragw}{HTML}{854F0B}   
\definecolor{rage}{HTML}{993556}   
\lstdefinestyle{ragsnippet}{language=json, basicstyle=\ttfamily\small,
  stringstyle=\color{black}, commentstyle=\color{gray}, frame=tb,
  columns=fullflexible, keepspaces=true, breaklines=true,showstringspaces=false,
  escapeinside={(*}{*)}, aboveskip=8pt, belowskip=8pt}
\newcommand{\cmark}{\ding{51}}  \newcommand{\xmark}{\ding{55}}
\newcommand{\pmark}{\tikz[baseline=-0.6ex]{\draw (0,0) circle (0.55ex); \fill (0,-0.55ex) arc (-90:90:0.55ex) -- cycle;}}
\lstdefinelanguage{json}{
  basicstyle=\ttfamily\footnotesize,
  morestring=[b]",
  stringstyle=\color{black},
  literate=
    {true}{{\textbf{true}}}4
    {false}{{\textbf{false}}}5
    {null}{{\textbf{null}}}4
}
\newcommand{\raglogv}[1]{RAGLog\textsubscript{#1}}

\newtcolorbox{answerbox}[1]{enhanced, breakable,
  colback=gray!8, colframe=gray!60!black,
  boxrule=0pt, leftrule=2.5pt, arc=0pt, outer arc=0pt,
  left=6pt, right=6pt, top=4pt, bottom=4pt,
  before skip=8pt, after skip=8pt,
  fontupper=\small,
  title={#1}, coltitle=black, fonttitle=\bfseries\small,
  attach title to upper={\ }}
\begin{document}

\title{From Logs to Faults: Reference-Free Fault Localization for Retrieval-Augmented Generation Pipelines}

\author{Hira Naveed}
\email{hira.naveed@monash.edu}
\orcid{0000-0002-0045-4577}
\correspondingauthor
\affiliation{%
  \institution{Monash University}
  \country{Australia}
}

\author{Scott Barnett}
\affiliation{%
  \institution{Deakin University}
  \country{Australia}}
\email{scott.barnett@deakin.edu.au}

\author{Stefanus Kurniawan}
\affiliation{%
  \institution{Deakin University}
  \country{Australia}
}

\author{John Grundy}
\affiliation{%
 \institution{Monash University}
 \country{Australia}}

\author{Chetan Arora}
\affiliation{%
  \institution{Monash University}
  \country{Australia}}

\author{Hourieh Khalajzadeh}
\affiliation{%
  \institution{Deakin University}
  \country{Australia}}

\author{Omar Haggag}
\affiliation{%
  \institution{Monash University}
  \country{Australia}}

\renewcommand{\shortauthors}{Naveed et al.}

\begin{abstract}
Retrieval-Augmented Generation (RAG) pipelines are widely used for domain-specific question answering, yet locating faults remains manual and time-consuming. Fault localization is challenging because RAG pipelines consist of interdependent stages, where faults can propagate downstream, masking their origin, or remain undetected even when the final answer is correct. Existing evaluation approaches often focus on individual stages and rely on reference answers or code instrumentation. We present RAGLog, which detects and localizes faults across RAG stages using execution logs, without requiring reference answers or code instrumentation. RAGLog transforms heterogeneous API logs into a common RAG schema and combines rule-based, embedding-based, and LLM-based checks to identify faulty stages. We introduce a live fault injector and RAGLog Faults, a dataset of 3,134 runs from five RAG pipelines with stage-level fault labels. With Claude Haiku 5.5, RAGLog achieves a pooled F1 score of 0.73 on runs with naturally occurring faults, outperforming Claude Code (0.58) and Codex (0.52), and comparable effectiveness on runs with injected faults (0.84 versus 0.80 and 0.82, respectively). It analyzes runs in approximately five seconds at US\$0.005 each, with overall lower cost and latency than agent baselines. RAGLog is available as an open-source tool for debugging, evaluating, and monitoring RAG pipelines.
\end{abstract}



 \received{31 October 2026}

\maketitle

\section{Introduction}\label{sec:intro}
Retrieval-Augmented Generation (RAG) is widely adopted for domain-specific question answering, as it enhances the
responses of large language models (LLMs) with query-relevant context retrieved from external knowledge sources
\cite{lewis2020retrieval, asai2023self, jeong2024adaptive, brehme2025retrieval}. It enables organizations to leverage
proprietary data without retraining models and reduces the risk of hallucination. While these advantages are
substantial, developing and maintaining RAG pipelines is not trivial. A RAG pipeline is not a single model but a sequence
of interdependent stages, typically comprising query rewriting, embedding, retrieval, reranking, and answer generation, many of which are realized through calls to third-party services such as hosted language models, embedding models, vector
databases, and reranking APIs \cite{barnett2024seven}.

Faults can arise at any of these stages, and their effects may propagate downstream, making it difficult to identify their origins \cite{barnett2024seven,lauro2025rag,sivasothy2025ragprobe}. For example, a rewriter may add a
constraint that the user never asked for, which leads the retriever to irrelevant chunks, and a generator may still
produce a correct answer from incomplete context.  Such effects are far from rare. In our study of 1,500 runs across five open-source RAG pipelines (Section~\ref{sec:dataset}), 37\% of the runs contained at least one faulty stage, and more than a third of these runs still produced a correct answer. Furthermore, among the 357 runs with incorrect answers, 71\% have at least one faulty stage before the generator. Detecting these faults is currently a manual and iterative process in which developers inspect logs and adjust parameters to debug RAG pipelines \cite{brehme2025retrieval, lauro2025rag}. This approach is error-prone and expensive in terms of time and resources. For example, it may require re-indexing all documents after parameter changes and then repeating the process several times \cite{lauro2025rag}.

Automated fault localization for RAG pipelines faces two main challenges. First, logs need to be converted into a form that can be analyzed. RAG pipelines are built from different providers, request formats, and call sequences; therefore, an automated approach must identify which call belongs to which stage, and what each stage received and returned, without manual configuration or code instrumentation. Second, faults need to be identified without a failure signal. Traditional fault localization starts from a known failure, such as a failing test, and searches for its cause \cite{wong2016survey}. However, a RAG pipeline provides no such signal, since reference answers are not available in production and not all faults change the final answer. Each stage therefore needs to be assessed on its own, without reference answers. Existing RAG evaluation frameworks measure how well documents are retrieved and how accurately responses are generated, but they often rely on ground truth and do not cover all stages of the pipeline \cite{es2024ragas, saad2024ares, deepeval2025rag, oleszak2025deepchecks}. Observability and tracing tools record each call but leave its analysis to the developer \cite{abedraba2025ragobservability, cheng2025ragtrace}, while general-purpose coding agents can analyze logs but are not designed to locate faults across RAG stages. These gaps highlight the need for an automated, reference-free approach to locate faults across the RAG pipeline.

In this study, we present RAGLog, a reference-free fault localization approach for debugging, evaluating, and monitoring RAG pipelines. RAGLog uses call-level execution logs, which contain one record for each API call that a pipeline stage makes to an external service, such as a language model, an embedding model, or a vector database. These logs can be collected through existing tracing tools, API gateways, or the logging proxy we provide with RAGLog. The approach first transforms execution logs into a common representation by classifying each call into a RAG stage, using rules for common APIs and an LLM for unfamiliar ones, and mapping each call to a common RAG schema. Then, a reference-free detector decides for every stage of every run if that stage is faulty. The detector combines rule-based and similarity-based measurements with targeted checks performed by a small LLM. A decision function learned from labeled runs combines these measurements into a verdict for each stage. Since the approach uses a common schema, a single detector works for multiple RAG pipelines.

 We evaluate RAGLog on five open-source RAG pipelines, which use models and services from seven providers, and three public datasets. In addition to 1,500 natural runs containing 670 faults, we evaluate RAGLog on 480 injected runs containing 419 faults. To evaluate RAGLog on faults with a known source, we developed a fault injector that modifies the input or output of a single stage while the pipeline runs, so that all later stages react to the fault as they would in practice. All development and evaluation runs, together with their stage-level fault labels and full traces, are released as RAGLog Faults, a dataset of 3,134 runs. 
 
To the best of our knowledge, no existing approach locates faults across all RAG pipeline stages without reference answers. Therefore, we compare RAGLog with two general-purpose coding agents, Claude Code and Codex, which are given the same logs, reflecting how a developer without a dedicated tool would approach the task. On natural runs, RAGLog, with Claude Haiku 5.5 as its underlying LLM, achieves a pooled F1 score of 0.73, compared to 0.58 for Claude Code and 0.52 for Codex, as the agents flag fewer stages and miss more than half of the faulty ones. On injected runs, the agents perform similarly, with pooled F1 scores of 0.80 and 0.82 compared to 0.84 for RAGLog. When an injected fault took effect, RAGLog flagged the correct stage in 90\% of the runs, compared to 80\% and 87\% for the agents. RAGLog also generalizes to previously unseen pipelines, outperforming both agents on natural runs, and automatically transforming logs of new pipelines into the common schema. Furthermore, RAGLog achieves these results at a fraction of Codex's cost and at a lower overall cost than Claude Code.

The main contributions of this work include:
\begin{itemize}
    \item \textbf{Reference-free, cross-stage fault localization.} RAGLog identifies faulty stages of a RAG pipeline using execution logs alone, without requiring reference answers or code instrumentation. By combining evidence across stages, it can detect faults that propagate to downstream stages or occur even when the final answer is correct. This is designed to support debugging failing runs and monitoring pipelines in production, where reference answers are often unavailable.
    \item \textbf{Infrastructure for observing, debugging, and evaluating RAG pipelines.} We provide a logging proxy, a RAG schema, and an automated mechanism for transforming execution logs from different providers into this schema without any manual configuration. We also introduce a live fault injector that modifies the input or output of a pipeline stage during execution, allowing later stages to respond to the injected fault as they would to a real fault. Together, these components enable developers to trace pipeline execution, evaluate how faults affect downstream stages, and assess changes in pipeline behavior, without modifying application code.
    
    \item \textbf{Empirical evidence on effectiveness, generalization, and cost.} We evaluate RAGLog across five open-source RAG pipelines and three public datasets, using 1,500 natural runs and 480 injected runs, and compare it against two general-purpose coding agents. The evaluation examines fault-localization effectiveness, generalization to unseen pipelines, the contribution of individual components, and analysis cost and time. The results also reveal that many injected faults are absorbed by later stages and that faulty stages do not necessarily lead to incorrect final answers.
    \item \textbf{Dataset and tool for debugging, evaluating, and monitoring RAG pipelines.} We release RAGLog Faults, a public dataset of 3,134 runs across five RAG pipelines, with stage-level fault labels, execution traces, and labeling judgments. We also provide an open-source RAGLog tool with a dashboard for debugging, evaluating, and monitoring RAG pipelines.
\end{itemize}

The remainder of this paper is structured as follows: Section~\ref{sec:motivation} presents a motivating example and introduces the stages of a RAG pipeline. Section~\ref{sec:approach} explains our approach, and Section~\ref{sec:dataset} describes the dataset. Section~\ref{sec:setup} describes the experimental setup, Section~\ref{sec:results} summarizes the experimental results, and Section~\ref{sec:tool} presents the RAGLog tool. Section~\ref{sec:discussion} discusses the implications of the findings, and Section \ref{sec:threats} discusses the threats to validity. Related work is reviewed in Section~\ref{sec:related}, followed by conclusions and directions for future work in Section~\ref{sec:conclusion}.

\section{Motivating Example} \label{sec:motivation}
\begin{figure*}[!htbp]
    \centering
    \includegraphics[width=0.9\linewidth]{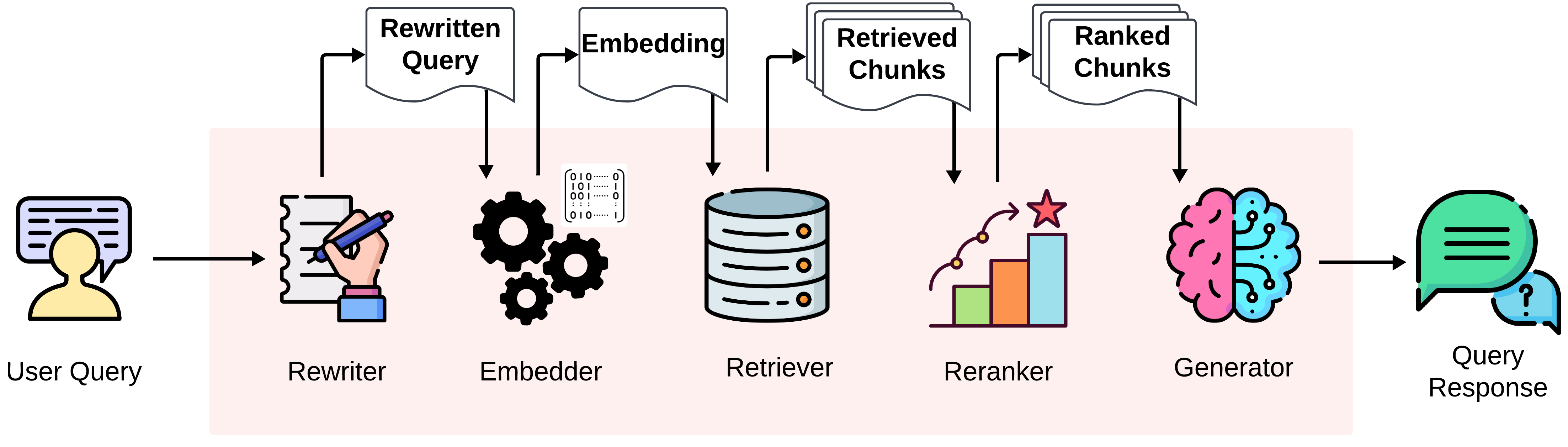}
    \caption{Stages of a typical RAG pipeline, such as Alex's.}
    \label{fig:pipeline}
\end{figure*}

Consider Alex, an AI engineer tasked with developing a question-answering chatbot for employees at a large enterprise. Alex decides to build a RAG pipeline since the questions will be based on a corpus of proprietary documents, and LLMs cannot correctly answer domain-specific questions based on their training data alone. The pipeline will take the user query as input, find query-relevant document chunks from the corpus, and produce a response based on the query and these chunks. To achieve this, Alex's pipeline will consist of the following stages: 1) a \textit{rewriter} that reformulates the user query to improve retrieval, 2) an \textit{embedder} that converts the query into a vector, 3) a \textit{retriever} that retrieves the chunks most similar to this vector from a vector database, 4) a \textit{reranker} that ranks the retrieved chunks by their relevance to the query and passes the top-ranked ones on, and
5) a \textit{generator} that produces the response from the query and the passed chunks. The rewriter and reranker are optional, and not every pipeline uses them. Alex builds the RAG pipeline shown in Figure \ref{fig:pipeline} by leveraging third-party APIs for all stages, as APIs scale well in production and do not require additional infrastructure, development, or maintenance overhead. 

To validate the pipeline, Alex asks several questions; while some return correct results, many others produce partially or completely incorrect responses. Alex realizes that the pipeline has several potentially faulty stages. We consider a stage faulty when its output deviates from the expected behavior for a given input, such as when the retriever fails to retrieve chunks containing the information needed to answer a question or when the generator produces claims that are not supported by the retrieved context. A fault in one stage may propagate downstream, so the stage where a wrong answer becomes visible is often not the stage that caused it. For example, an employee asks
\textit{``How many days of parental leave are new employees entitled to?''}, and the chatbot replies that the documents
do not contain this information. Alex suspects the retriever, but the logs show that the generator received a chunk of
the leave policy stating the number of days and ignored it; the fault lies in the generator, not in retrieval. In another
case, an employee asks \textit{``What does GDPR stand for?''}, and the chatbot answers correctly. However, the retriever
did not return the section of the document that defines the term, and the generator answered from its own
training data instead. The answer is correct, but the retriever's fault remains hidden, and it will affect any question
whose answer the generator cannot know, such as questions about the company's own procedures.
Locating such faults requires inspecting every stage of every run, which Alex can only do by reading the logs of each
API call, a laborious and error-prone process.

The absence of reference answers further complicates fault localization. Before deploying the chatbot, Alex wants to
evaluate it on questions from the company documents. This requires domain experts to write reference answers, which is expensive and must be repeated whenever the documents change
\cite{zhu2025rageval}. Although synthetic test sets can reduce this cost \cite{es2024ragas}, they still require validation from domain experts. Even if such domain-specific evaluation is conducted, in production the chatbot will answer real employee queries, for which reference answers will not be available. Alex's RAG pipeline is built using third-party APIs, so internal instrumentation is also not possible. As a result, the only way she can learn about production faults is through log inspection or user reports of incorrect answers. 

\begin{table}[!htpb]
\caption{Existing approaches for detecting and locating faults in RAG pipelines.}
\label{tab:existing}
\small
\begin{tabular}{lcccccc}
\toprule
Approach & \shortstack{All RAG\\stages} & \shortstack{Locates the\\faulty stage} & \shortstack{No reference\\answers needed} & \shortstack{No known\\failure needed} & \shortstack{No code\\changes needed} &  \shortstack{Fully\\Automated} \\
\midrule
RAGAS \cite{es2024ragas}                            & \xmark & \xmark & \pmark & \cmark & \xmark & \cmark \\ 
ARES \cite{saad2024ares}                            & \xmark & \xmark & \pmark & \cmark & \xmark & \cmark \\ 
RAGChecker \cite{ru2024ragchecker}                  & \xmark & \pmark & \xmark & \cmark & \xmark & \cmark \\ 
RAGEval \cite{zhu2025rageval}                       & \xmark & \xmark & \pmark & \cmark & \xmark & \cmark \\ 
RAGProbe \cite{sivasothy2025ragprobe}               & \xmark & \xmark & \pmark & \cmark & \cmark & \cmark \\ 
RAGEC classifier \cite{leung2026classifying}              & \pmark & \cmark & \xmark & \pmark & \xmark & \cmark \\ 
RAG-Critic \cite{dong2025rag}                 & \xmark & \pmark & \cmark & \cmark & \xmark & \cmark \\ 
Doctor-RAG \cite{jiao2026doctor}                 & \pmark & \cmark & \cmark & \xmark & \xmark & \cmark \\ 
MetaRAG \cite{sok2025metarag}                       & \xmark & \xmark & \cmark & \cmark & \pmark & \cmark \\ 
Raggy \cite{lauro2025rag}                           & \pmark & \xmark & \cmark & \cmark & \xmark & \xmark \\ 
RAGTrace \cite{cheng2025ragtrace}                   & \pmark & \xmark & \cmark & \cmark & \xmark & \xmark \\ 
Langfuse \cite{langfuse}                    & \cmark & \xmark & \cmark & \cmark & \pmark & \pmark \\ 
LangSmith \cite{langsmith}                  & \cmark & \xmark & \cmark & \cmark & \pmark & \pmark \\ 
Arize Phoenix \cite{phoenix}                & \cmark & \xmark & \cmark & \cmark & \pmark & \pmark \\ 
\bottomrule
\end{tabular}
\end{table}

  Therefore, she starts exploring existing solutions that can locate faults in RAG pipelines, summarized in Table~\ref{tab:existing}. RAG evaluation frameworks \cite{es2024ragas, saad2024ares, ru2024ragchecker} score retrieval and generation quality, but do not cover the other stages, rely on reference answers for several of their metrics, and do not determine which stage is at fault. Test-generation approaches \cite{zhu2025rageval, sivasothy2025ragprobe} reduce the cost of reference answers by generating question-answer pairs from the documents, which reveal pipeline failures but do not identify the root cause. Recent diagnostic approaches go further and attribute faults to stages of the RAG pipeline, yet they all have certain limitations: RAGEC classifier \cite{leung2026classifying} requires reference answers, RAG-Critic \cite{dong2025rag} distinguishes only retrieval from generation errors in responses it judges as incorrect, and Doctor-RAG \cite{jiao2026doctor} analyzes only agentic runs already known to have failed. Hallucination detectors such as MetaRAG \cite{sok2025metarag} work without reference answers, but examine only the generator. Interactive debugging tools \cite{lauro2025rag, cheng2025ragtrace} make the behavior of a pipeline visible, but leave its analysis to the developer. Observability platforms \cite{langfuse, langsmith, phoenix} trace every stage of a pipeline and offer automated evaluators for retrieval and generation quality, but they score stages individually rather than determining which stage caused a fault, and they usually require the pipeline's code to be instrumented. Many of these approaches rely on large LLM judges or several LLM calls per answer, which makes analyzing all pipeline runs costly. 

This motivating example highlights the following features that Alex requires for locating faults in her RAG pipeline:
\begin{itemize}
    \item Automated fault localization across all stages of the RAG pipeline, including faults that do not change the final answer.
    
    \item Independence from reference answers, enabling pipeline evaluation on domain-specific questions and monitoring in production environments.
    
    \item Integration with RAG pipelines without any changes to the source code.
    
    \item Low cost and resource usage, enabling every run to be analyzed in production.
\end{itemize}

\section{Approach}\label{sec:approach}
\begin{figure*}[!htbp]
  \includegraphics[width=\textwidth]{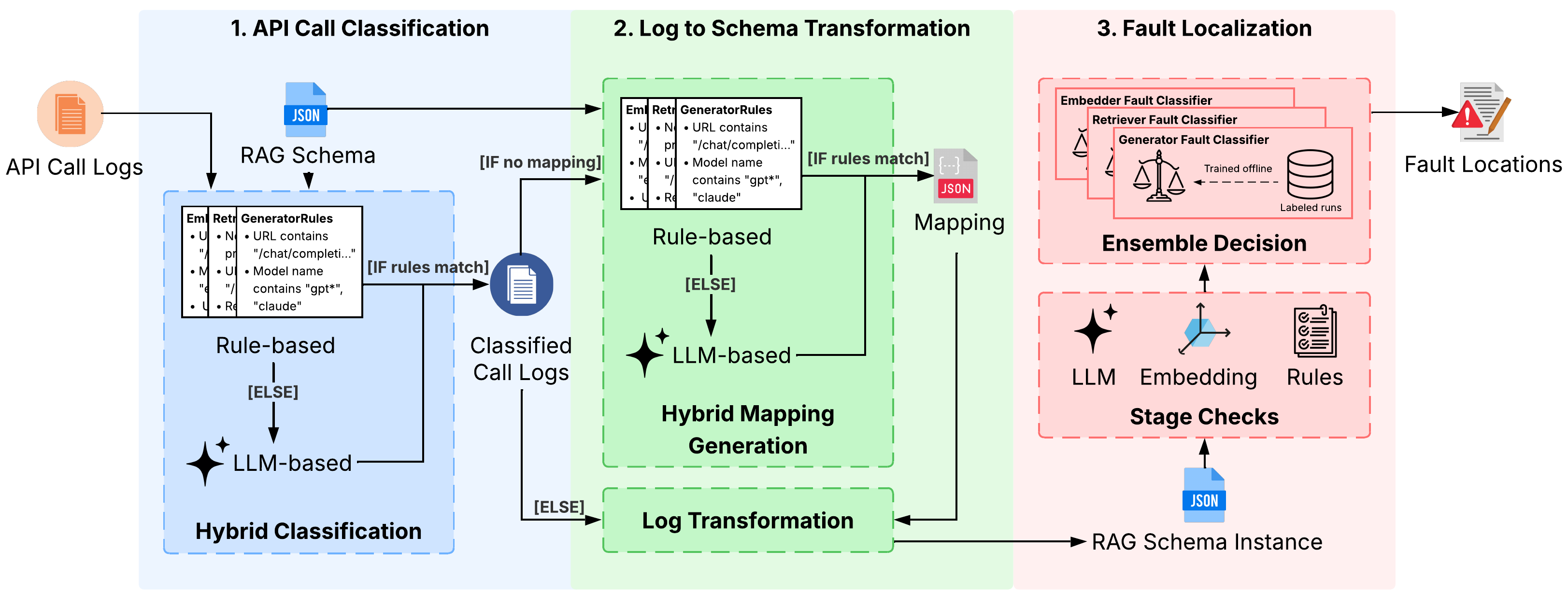}
  \caption{RAGLog: Approach Overview}
  \label{fig:approach}
\end{figure*}

To address the challenges described in Section~\ref{sec:motivation}, we present RAGLog, a reference-free fault localization approach for RAG pipelines. RAGLog takes the execution logs of a RAG pipeline as input and decides, for every stage of every run, whether that stage is faulty. Throughout this paper, \textit{fault localization} covers both detecting that a run contains a fault and identifying the stage that is faulty. Figure~\ref{fig:approach} shows an
overview of RAGLog, which consists of three components. First, each logged API call is classified into the RAG stage defined in our schema. Second, the classified calls are transformed into instances of the RAG schema, one instance per run, using a mapping that is generated once for each pipeline. Third, stage checks based on an LLM, an embedding model, and rules measure each stage in the schema instance, and an ensemble decision of stage-specific fault classifiers turns these measurements into the fault locations of the run. 
We first describe the RAG schema and then each of the components in detail below.

\subsection{RAG Schema} \label{sec:schema}
\begin{table}[!htbp]
    \centering
    \caption{RAG schema structure}
    \label{tab:schema}
    \small
    \begin{tabular}{lll}
        \toprule
        Stage & Inputs & Outputs \\
        \midrule
        Rewriter (optional) &  System prompt \& user query & Refined user query \\
        Embedder & User query & Query embedding \\
        Retriever &  Query embedding & Retrieved chunks \\
        Reranker (optional) &  Query \& retrieved chunks & Reranked chunks  \\
        Generator &  System prompt, user query \& retrieved chunks &
            Query response \\
        \bottomrule
    \end{tabular}
\end{table}

The RAG schema serves as a template that captures the architecture of a typical RAG pipeline, including its stages and the inputs and outputs of each stage. The schema is populated using execution logs, with each instance representing a single run including all API calls made by the pipeline to answer one question. These populated instances are later used for fault localization. We include five stages, shown in Table \ref{tab:schema}: 1) rewriter, 2) embedder, 3) retriever, 4) reranker, and 5) generator: 

\begin{itemize}
\item \textit{Rewriter}: reformulates the user query to improve its effectiveness in subsequent stages. It takes a system prompt describing the task and the user query, and produces a refined query that is optimized for downstream stages. The rewriter is optional.

\item \textit{Embedder}: converts the user query into a semantic vector representation for retrieval from a vector database. It takes the user query, or the refined query when a rewriter is present, and produces the corresponding query embedding.

\item \textit{Retriever}: performs a similarity search over the vector database using the query embedding produced by the embedder. It returns the retrieved chunks, a ranked set of document chunks that may contain query-relevant information.

\item \textit{Reranker}: refines the ordering of the retrieved chunks by estimating their relevance to the query. It
takes the query together with the retrieved chunks and produces the reranked chunks, from which the most relevant ones are passed on to the generator. This stage is also optional. 

\item \textit{Generator}: uses an LLM to answer the user query based on the retrieved chunks. It takes a system prompt containing generation instructions, the user query, and the retrieved chunks (or reranked chunks, if a reranker is present), and produces the final response.

\end{itemize}

\subsection{API Call Classification} \label{sec:classification}
This component classifies raw API call logs according to RAG pipeline stages defined in our schema (Section \ref{sec:schema}). Since raw logs do not explicitly identify the stage associated with each API call, they cannot be used directly for fault localization. We first pair corresponding request and response logs using API call IDs, and then classify each pair with a hybrid strategy that combines efficient rule-based classification for known APIs with an LLM fallback for unseen ones. The classification also filters out irrelevant API calls, such as health checks. An example is shown in Listing \ref{lst:rag-classification}.

\begin{lstlisting}[style=ragsnippet, caption={A snippet of a reranker call logged before and after classification}, label={lst:rag-classification}]
{"id": "4f2c91",
 "method": "POST",
 "url": (*"https://api.cohere.com/v2\textcolor{ragx}{/rerank}"*),
 "model_name": "rerank-english-v3.0",
 "raw_request":  {(*\textcolor{ragw}{"query": "parental leave days new employees"}*),
                  (*\textcolor{ragc}{"documents": ["Annual leave accrues ...",}*)
                               (*\textcolor{ragc}{"New employees are entitled to 90 days ...", ...]}*)},
 "raw_response": {(*\textcolor{ragr}{"results": [\{"index": 1, "relevance\_score": 0.93\},}*)
                              (*\textcolor{ragr}{\{"index": 0, "relevance\_score": 0.41\}, ...]}*)}}

// The same call after classification
{"id": "4f2c91", ...,
 (*\textcolor{ragx}{"stage": "reranker",}*)
 (*\textcolor{ragx}{"classifier\_used": "rule",}*)
 (*\textcolor{ragx}{"classification\_reasoning": "RULE: URL matches: /rerank"}*)}
\end{lstlisting}

\subsubsection{Rule-based Classification}
The logs are first processed by a rule-based classifier that classifies each request-response pair as one of the five stages defined in our schema. The classification is performed by identifying patterns such as keywords, URLs, model names, and request and response formats. For instance, we detect embedder calls by examining the model name and URL for embedding-related keywords such as 'embedding', 'embed', 'encode', and 'text-embedding'. The classifier also considers the position and content of each call within a run to distinguish between similar stages, such as the rewriter and generator, which typically use the same LLM endpoint. While the rule-based classifier can detect most stages, it is not robust to API changes or unseen patterns. When a request-response pair does not match any predefined rule, classification falls back to the LLM-based classifier.

\subsubsection{LLM-based Classification}
The LLM-based classifier sequentially performs binary classification on the request-response pair for each of the five stages. Each prompt provided to the LLM describes the characteristics of the stage, and contains the URL, model name, and request-response pair. The prompt also includes positive and negative examples to improve classification performance. Since the order of calls carries information, the prompt also contains the stage of the previous call. Once the LLM identifies a stage, the remaining stages are not checked. If a request–response pair matches no stage in either classifier, it is marked as irrelevant. We use Claude Haiku 5.5 for API call classification.

\subsection{Log-to-Schema Transformation}\label{sec:transformation}
This component transforms the classified call logs from the previous step into instances of the RAG schema, providing a structured format that can later be used for fault localization. A mapping generator identifies how fields in each call map to fields in the RAG schema, then uses this mapping to transform the calls into schema instances. The mapping is created only once for a given RAG pipeline; subsequent calls are transformed into schema instances using the existing mappings. Similar to API call classification, we use a hybrid strategy for mapping generation with a rule-based generation followed by an LLM fallback. Listing \ref{lst:rag-mapping} shows an example of rule-based mapping generation.

\begin{lstlisting}[style=ragsnippet, caption={Mapping generation for the call in Listing~\ref{lst:rag-classification}: a predefined rule matches the call and becomes the mapping for the reranker}, label={lst:rag-mapping}]
// Predefined rule for Cohere's rerank format
{"stage": "reranker",
 "inputs":  [{"input_key": (*\textcolor{ragw}{"query"}*),      "target": (*\textcolor{ragw}{"userQuery"}*)},
             {"input_key": (*\textcolor{ragc}{"documents"}*),  "target": (*\textcolor{ragc}{"retrievedChunks"}*)}],
 "outputs": [{"output_key": (*\textcolor{ragr}{"results[*]"}*), "target": (*\textcolor{ragr}{"rerankedChunks"}*)}]}
// query, documents and results exist in the call: the rule matches

// Mapping stored for this RAG pipeline
{"stage": "reranker",
 "inputs":  {(*\textcolor{ragw}{"userQuery": "<query>"}*),
             (*\textcolor{ragc}{"retrievedChunks": ["<documents>"]}*)},
 "outputs": {(*\textcolor{ragr}{"rerankedChunks": "<results[*]>"}*)}}
\end{lstlisting}
\subsubsection{Rule-based Mapping Generation} A mapping describes which fields of the request and response correspond to the inputs
and outputs of that stage in the schema. For example, the mapping of a reranker call maps the \textit{query} and
\textit{documents} fields of the request to the query and retrieved chunks of the schema, and the \textit{results}
field of the response to the reranked chunks in the schema. For each stage in the schema, we define several such rules to support different API providers. If a mapping does not exist for a RAG pipeline, the rule-based mapping generator creates one by matching the calls from each stage against predefined rules for common API formats. For example, a call classified as the embedder stage would be matched with mapping rules for common embedding providers such as OpenAI, Nomic AI, Cohere, and BGE. The rules are extensive enough to cover most log formats; however, if a match is not found, the mapping generation is handled by the LLM-based mapping generator. 

\subsubsection{LLM-based Mapping Generation} We prompt an LLM to generate mappings between a classified call and its corresponding schema subset. The prompt contains
the request and response of the call, the schema fields of the stage, a description of the mapping task, and positive and negative examples. The LLM returns, for every schema field, the path of the corresponding field in the call. The mapping is stored as a JSON file for each pipeline. We use Claude Haiku 5.5 for mapping generation. 

\subsubsection{Log Transformation} This step uses the created mappings to transform classified call logs into schema instances. A JSON-to-JSON mapper applies the mapping to each call, extracts the values at the mapped paths, and populates the corresponding fields of the schema instance. An example is shown in Listing \ref{lst:rag-instance}. RAGLog then groups the mapped calls into individual runs, each representing a single question and ending with the generator call that produces the answer. Each run forms a separate schema instance used for fault localization. To account for stages that may not generate API calls, RAGLog reconstructs missing information where possible. For example, some pipelines use an in-process vector database for retrieval, leaving no retriever API call in the logs. In such cases, the retriever's output is reconstructed from the reranker's input, which contains the retrieved chunks in their original retrieval order, allowing the retriever to be assessed even without an API call. 

\begin{lstlisting}[style=ragsnippet, caption={
RAG schema instance for the reranker stage generated by transforming the API call in Listing~\ref{lst:rag-classification} using the mapping in Listing~\ref{lst:rag-mapping}}, label={lst:rag-instance}]
{"stage": "reranker",
 "inputs":  {(*\textcolor{ragw}{"userQuery": "parental leave days new employees"}*),
             (*\textcolor{ragc}{"retrievedChunks": ["Annual leave accrues ...",}*)
                             (*\textcolor{ragc}{"New employees are entitled to 90 days ...", ...]}*)},
 "outputs": {(*\textcolor{ragr}{"rerankedChunks": [\{"index": 1, "relevance\_score": 0.93\},}*)
                               (*\textcolor{ragr}{\{"index": 0, "relevance\_score": 0.41\}, ...]}*)}}
\end{lstlisting}
\subsection{Fault Localization} \label{sec:localization}
This component locates faults in a RAG pipeline using a populated instance of the RAG schema. It determines whether each stage in the instance is faulty by examining its inputs and outputs, as well as those of other stages. Without reference answers, this requires examining how information flows through the pipeline. For example, whether the retrieved chunks contain the information needed to answer the question, whether the reranker preserves that information, and whether the generator uses it to produce the answer. A capable reasoning model such as Opus 5.5 \cite{anthropic2026opus55} could perform this analysis, but its cost makes it impractical for analyzing every run in production. Smaller LLMs, such as GPT-5 Mini \cite{openai2025gpt5mini}, are less reliable when asked to identify faulty stages directly. Similarly, embedding similarities and rules are computationally inexpensive but can only detect certain types of faults. RAGLog therefore divides fault localization into two parts. First, three types of checks assess specific properties of each stage: rule-based checks, embedding-based checks, and LLM-based checks. Second, an ensemble decision combines the measurements collected across the entire run to determine which stage is faulty using a fault classifier trained on labeled runs for each stage.

\subsubsection{Stage Checks} \label{sec:checks}
Each stage check produces one or more numeric measurements, such as a similarity, a count, or a confidence. Table~\ref{tab:checks} lists the checks and the stages they measure.
\begin{table}[!htbp]
    \centering
    \caption{Stage checks of RAGLog}
    \label{tab:checks}
    \small
    \setlength{\tabcolsep}{4pt}
    \begin{tabular}{>{\raggedright\arraybackslash}p{2.4cm}>{\raggedright\arraybackslash}p{1.2cm}>{\raggedright\arraybackslash}p{7.4cm}>{\raggedright\arraybackslash}p{3.0cm}}
        \toprule
        Check & Stage & What it measures & Output \\
        \midrule
        \multicolumn{4}{l}{\textit{Rule-based}} \\
        Rewriter rules & Rewriter & Question's names, numbers, and dates missing from every rewrite; names and
            numbers added by the rewrites; content words lost; reversed intent; changed question type; number of
            rewrites & Shares, counts, and flags \\
        \cmidrule(l){2-4}
        Embedder rules & Embedder & Question's names missing from the embedder input; misspelled words; added words;
            length relative to the question; corrupted text & Shares, counts, ratio, and flag \\
        \midrule
        \multicolumn{4}{l}{\textit{Embedding-based}} \\
        Query drift & Rewriter & Similarity of each rewrite to the question & Similarities \\
                    & Embedder & Similarity of the embedder input to the question and its rewrites & Similarities \\
        \cmidrule(l){2-4}
        Retrieval coverage & Retriever & Similarity of the best retrieved chunk to the question; rank of the most
            similar chunk; agreement of the retrieval order with the similarity order & Similarity, relative rank,
            rank correlation \\
        \cmidrule(l){2-4}
        Reranking order & Reranker & Agreement of the reranked order with the similarity order; position of the most
            similar chunk; similarity gap between the top-ranked and the most similar chunk & Rank correlation,
            relative position, similarity gap \\
        \cmidrule(l){2-4}
        Answer grounding & Generator & Similarity of the answer to its context; answer given with an empty context;
            refusal despite a relevant chunk & Similarities and flag \\
        \midrule
        \multicolumn{4}{l}{\textit{LLM-based}} \\
        Graded stage check & Rewriter & Three fault types: a rewrite drops a constraint, adds a constraint, or
            changes an entity & Confidence (0--100) per fault type \\
                           & Embedder & One fault type: the embedder input is not a faithful rendering of the
            question & Confidence (0--100) \\
                           & Retriever & Three fault types: no retrieved chunk answers the question, the chunks
            cover only part of it, or most chunks are irrelevant & Confidence (0--100) per fault type \\
                           & Reranker & Two fault types: a good chunk was excluded, or weak chunks were passed on &
            Confidence (0--100) per fault type \\
        \cmidrule(l){2-4}
        Graded generator check & Generator & Six fault types: the answer is unsupported, off the question,
            incomplete, should have abstained, wrongly refuses, or misstates a number & Confidence (0--100) per fault
            type \\
        \cmidrule(l){2-4}
        Generator verification & Generator & Six fault types defined by explicit rules, such as information about a
            different entity or a missing item that the question asks for, and any fault & Confidence (0--100) per
            fault type \\
        \cmidrule(l){2-4}
        Evidence flow check & Retriever & Pieces of information the question needs that no passed or excluded chunk
            contains, or contains only partly & Yes, partly, or no per piece; shares of pieces \\
                            & Reranker & Needed pieces that only chunks excluded by the reranker contain & Yes,
            partly, or no per piece; shares of pieces \\
                            & Generator & Needed pieces that were passed on but that the answer omits, states
            differently, or says are unknown; the answer's unsupported statements; refusal & Answer treatment per
            piece; shares and counts \\
        \cmidrule(l){2-4}
        Statement-level generator check & Generator & Statements of the answer that its context does not state,
            classified by materiality; omitted asked items, wrong refusal, answer to a different question & Count per
            category; confidence (0--100) per fault type \\
        \bottomrule
    \end{tabular}
\end{table}

Together, these checks target the fault types identified in the RAG error taxonomy RAGEC \cite{leung2026classifying} that are observable in RAG pipeline execution logs. These faults are also documented in other studies, including: failures to retrieve relevant chunks or the retrieval of irrelevant chunks \cite{barnett2024seven, lauro2025rag, dong2025rag, ru2024ragchecker, jiao2026doctor}; relevant chunks being ranked too low to reach the generator or being obscured by irrelevant chunks \cite{barnett2024seven, lauro2025rag, cuconasu2024power, liu2024lost}; generated answers containing claims unsupported by or inconsistent with the retrieved context \cite{niu2024ragtruth, ru2024ragchecker, dong2025rag, sok2025metarag}; and answers that are incomplete, fail to address the question, or disregard the document context, such as answering when the available context is insufficient and the system should abstain \cite{barnett2024seven, lauro2025rag, dong2025rag, chen2024benchmarking, joren2025sufficient}. In addition, the checks target two fault types not covered by RAGEC: query rewrites that alter the original question \cite{ma2023query, li2024dmqr, jiao2026doctor} and embeddings that fail to faithfully represent the query text. Some checks compare a stage’s output with the original question, so they can also fire when an earlier stage caused the problem; the fault classifiers combine the checks of all stages to decide which stage is responsible.

\noindent\textit{\textbf{Rule-based checks.}}
The rule-based checks assess the rewriter and embedder, where faults often involve individual words, such as a missing year or a misspelled name, that embedding similarities may fail to detect. For the rewriter, the checks extract names, numbers, and dates from the original question and measure how many are missing from the rewrites, as well as any names or numbers introduced by the rewrites. The checks also identify changes in intent, such as an added negation, and changes in question type, such as a \textit{how} question becoming a \textit{what} question. For the embedder, the checks compare the embedder input with the original question and its rewrites to identify missing names, words differing by one or two edits from the original, added words, changes in relative text length, and corrupted text.

\noindent\textit{\textbf{Embedding-based checks.}}
The embedding-based checks use a small local embedding model (all-MiniLM-L6-v2 \cite{reimers2019sentence, sentenceTransformersMiniLM}) to measure semantic similarities across pipeline stages. For the rewriter and embedder, they measure how much the rewrites and embedder input deviate from the original question. For the retriever and reranker, they assess the similarity between the question and the retrieved chunks, the rank of the most similar chunk, and how closely the retrieval and reranking orders align with question–chunk similarity, using Spearman's rank correlation. For the generator, the checks measure the similarity between the generated answer and its passed document chunks, detect whether an answer was produced with an empty context, and identify cases where the generator refuses to answer despite having a relevant chunk in its context.

\noindent\textit{\textbf{LLM-based checks.}}
Many faults can only be recognized by understanding the question, the chunks, and the answer. For example, a retrieved chunk may refer to a different person with the same name, or an answer may introduce a date that is not supported by any chunk. RAGLog uses five targeted LLM-based checks to identify such faults. Each check asks a small LLM (by default Claude Haiku 5.5; Section~\ref{sec:setup-raglog} evaluates variants of RAGLog with two other LLMs) to assess specific aspects of a run using only the available execution data, without relying on external knowledge or reference answers. The LLM returns its findings in a predefined JSON format. Most checks request confidence scores between 0 and 100 rather than binary judgments, allowing the ensemble decision mechanism to account for uncertainty.

\begin{itemize}
    \item The \textit{graded stage check} examines the user query, rewritten query, embedder input, retrieved chunks in their final order, and the chunks passed to the generator. It requests confidence scores for nine fault types associated with the rewriter, embedder, retriever, and reranker stages, as listed in Table~\ref{tab:checks}. Stages that are absent from the pipeline are explicitly marked as not present.
    
    \item The \textit{graded generator check} examines the user query, the chunks passed to the generator, and the generated answer. It requests confidence scores for six generator fault types, including unsupported claims and incorrect refusals.
    
    \item The \textit{generator verification} assesses similar generator fault types but provides explicit criteria for each. For example, leaving out optional details does not make an answer incomplete, while information concerning a different entity, case, or time period cannot be used to support the answer.
    
    \item The \textit{evidence flow check} examines how information passes through the retriever, reranker, and generator. The LLM first identifies the individual pieces of information needed to answer the question. For each piece, it determines whether the information appears in the chunks passed to the generator or in retrieved chunks excluded by the reranker, using \textit{yes}, \textit{partly}, or \textit{no} judgments. It also determines whether the generated answer includes the information, omits it, states something different, or indicates that it is unknown. RAGLog uses these judgments to derive stage-specific measurements, such as the proportion of required information missing from all retrieved chunks. Information absent from the retrieved chunks indicates a potential retriever fault; information present only in excluded chunks indicates a potential reranker fault; and information passed to the generator but omitted or misstated in the answer indicates a potential generator fault. When many chunks are retrieved, the check includes all chunks passed to the generator and a shortlist of excluded chunks most relevant to the question, selected using a local cross-encoder (ms-marco-MiniLM-L-6-v2 \cite{reimers2019sentence, crossEncoderMsMarco}).
    
    \item The \textit{statement-level generator check} identifies statements in the generated answer that are not supported by the provided chunks and classifies them into categories, such as contradictions, unsupported answers to the requested facts, added names or numbers, and harmless elaborations or hedged remarks. Only the first three categories are considered faults, avoiding false alarms for paraphrasing or general explanations. RAGLog records the number of statements in each category. The LLM also gives a confidence for each of four fault types: the answer states a fact that the chunks do not support, omits an item that the question asks for, wrongly refuses, or answers a different question.
\end{itemize}
The five LLM-based checks run independently and are executed in parallel with the rule-based and embedding-based checks. As a result, the overall analysis latency is primarily determined by the slowest LLM call.

\subsubsection{Ensemble Decision}\label{sec:ensemble}
The ensemble decision turns the stage check measurements of a run into a verdict for each stage. Using fixed rules for this process has two limitations. First, measurements vary in scale and reliability, and their importance may differ across stages. Second, evidence of a fault in one stage may appear in the inputs or outputs of another stage. Hence, RAGLog uses a separate fault classifier for each stage, which uses measurements from all stages of the run to estimate the probability that the stage is faulty.

All fault classifiers combine three types of models: logistic regression, which provides a simple regularized model when relatively few faulty examples are available; one gradient-boosted tree model \cite{ke2017lightgbm}, which captures interactions between measurements; and an extremely randomized trees model \cite{geurts2006extremely}, whose randomized splits give smoother,
lower-variance probabilities. Their predictions are combined using a weighted average to estimate the probability that stage $s$ is faulty given the measurements $x$:

\begin{equation}
     p_s(x) = \tfrac{1}{3}\,p^{\mathrm{LR}}_s(x) + \tfrac{1}{2}\,p^{\mathrm{GB}}_s(x) + \tfrac{1}{6}\,p^{\mathrm{ET}}_s(x),
\end{equation}

where logistic regression contributes one third of the prediction, the gradient-boosted model contributes one half, and extremely randomized trees contribute one sixth. The ensemble weights were fixed during development, and alternatives, including equal weights, changed the pooled F1 score by no more than approximately 0.03. 

To account for the lower frequency of faulty stages, all models use class weights inversely proportional to class frequency. Missing measurements are handled natively by the gradient-boosted models and replaced with median values for the other models. To limit overfitting given the relatively small number of faulty runs per stage in the training data, we use regularized logistic regression ($C=0.5$), shallow trees (maximum depth of three, 100 iterations, a learning rate of 0.05, and at least ten runs per leaf) and 300 extremely randomized trees with at least three runs per leaf.

A stage is classified as faulty when its predicted probability $p_s(x)$ reaches a stage-specific threshold $\tau_s$. To limit false alarms, we select the threshold that maximizes the F1 score for each stage while keeping its false-positive rate at or below 10\%. Thresholds are selected using out-of-fold predictions rather than predictions on the data used to train the models. Specifically, the training runs are divided into three folds across three different partitions, with all runs associated with the same question kept in the same fold. Each stage is assessed only in runs where that stage is present. Finally, RAGLog clears a retriever flag when the evidence flow check finds the needed information only in chunks that the reranker excluded, attributing the fault to the reranker instead.

The stage-specific fault classifiers and their thresholds are trained once, offline, using runs with stage-level fault labels. Since all measurements are derived from the RAG schema instance, RAGLog can be applied to new pipelines without additional labeling or training. In this study, we train the classifiers using the development runs of the RAGLog Faults dataset (Section~\ref{sec:dataset}). For each stage in a run, RAGLog reports its fault classification, predicted probability $p_s(x)$, and plain-language explanations derived from the LLM-based checks.

\section{The RAGLog Faults Dataset}\label{sec:dataset}
Evaluating fault localization in RAG pipelines requires runs of real pipelines in which every stage is labeled as faulty or not. Existing RAG benchmarks provide questions, documents, and reference answers, but no labels for the stages of a
pipeline, and existing studies of RAG failures describe faults without releasing the runs in which they occurred
\cite{barnett2024seven, lauro2025rag}. We therefore built RAGLog Faults, a dataset of 3,134 runs of five open-source RAG pipelines on three public question-answering datasets, in which every observable stage of every run is labeled. Each run contains the question, its reference answer and evidence, the answer of the pipeline, the full trace of the run, the stage-level fault labels with their fault types, and the judgments from which the labels were derived. The dataset contains natural runs with faults that the pipelines produced on their own, and injected runs, in which we introduced faults into individual stages while the pipelines were running. The runs are divided into development and test sets, with no questions shared between them. The development runs were used to develop and train RAGLog, while the test runs were used exclusively for evaluation. This section describes the dataset construction and characteristics in detail.

\subsection{RAG Pipelines and Source Datasets}\label{sec:dataset-pipelines}
We use five open-source RAG pipelines: RAGFlow \cite{ragflow}, LightRAG \cite{lightrag}, Quivr \cite{quivr}, Onyx \cite{onyx}, and Verba \cite{verba}. 
These pipelines were selected based on three criteria: community adoption as measured by GitHub stars, architectural diversity covering different combinations of stages and models, and support for container deployment with Docker \cite{docker}. Table~\ref{tab:pipelines} lists the stages of each pipeline and the models used. Two pipelines perform query rewriting: Quivr refines the user query, while Onyx generates multiple keyword queries. Three pipelines, RAGFlow, LightRAG, and Quivr, rerank the retrieved chunks. All five pipelines embed the query, retrieve relevant chunks, and generate answers. However, RAGFlow and Quivr search their vector databases in-process, without making retriever API calls. For these two pipelines, the retrieved chunks are reconstructed from the input of the reranker (Section~\ref{sec:transformation}).

\begin{table}[!htbp]
    \centering
    \caption{Stages of the five RAG pipelines}
    \label{tab:pipelines}
    \small
    \setlength{\tabcolsep}{4pt}
    \begin{tabular}{@{}llllll@{}}
        \toprule
        Pipeline & Rewriter & Embedder & Retriever & Reranker & Generator \\
        \midrule
        RAGFlow & -- & Cohere embed-english-v3.0 & in-process$^{\ast}$ & Cohere rerank-english-v3.0 & GPT-4.1 nano \\
        LightRAG & -- & BGE-M3 & Qdrant & Jina reranker v2 & GPT-4o \\
        Quivr & GPT-4o & OpenAI text-embedding-ada-002 & in-process$^{\ast}$ & Cohere rerank-multilingual-v3.0 & GPT-4o \\
        Onyx & Gemini 3.5 Flash & Nomic embed-text-v1 & Vespa & -- & Gemini 3.5 Flash \\
        Verba & -- & Nomic embed-text & Weaviate & -- & Llama 3 \\
        \bottomrule
    \end{tabular}

    \vspace{1mm}
   \footnotesize $^{\ast}$ Retrieval is performed in-process; retrieved chunks are reconstructed from the reranker's input.\\ --: Stage absent.
\end{table}

To make the pipelines comparable, we harmonized their settings where their design allowed: chunks of about 512 tokens with an overlap of 50 tokens, six chunks passed to the generator, up to 64 candidates for the reranker, and no chat history between questions.

We use three public question-answering datasets with documents, questions, reference answers, and their evidence from different domains. DragonBall, part of the RAGEval benchmark \cite{zhu2025rageval}, contains generated documents from finance, law, and medical domains with questions of several types, such as factual, multi-hop, and summary questions. Natural Questions (NQ) \cite{kwiatkowski2019natural} contains real questions asked on the Google search engine, each paired with a Wikipedia article. MS MARCO \cite{bajaj2016ms} contains real questions asked on the Bing search engine with web passages. For DragonBall, we use its entire English corpus of 108 documents, over which all of its questions are asked. For NQ and MS MARCO, we built a corpus of 200 documents each that contains the documents referenced by the sampled questions. From all three datasets, we sampled 150 questions, 50 for development and 100 for testing, with no overlap between the two; the DragonBall test questions are stratified by question type and domain.

\subsection{Collecting Execution Logs}\label{sec:dataset-logs}
To collect execution logs without changing the code of the RAG pipelines, we built a logging proxy based on mitmproxy
\cite{mitmproxy}, an interactive man-in-the-middle proxy that can record and modify HTTP and HTTPS traffic. The proxy
is added as a service to the Docker Compose \cite{docker_compose} file of a pipeline, and the pipeline sends its
outbound requests through it by setting the standard proxy environment variables and trusting the proxy's certificate.
For every API call, the proxy records the URL, the HTTP method, the request and response content, the status code, and
timestamps. A second instance of the proxy runs in front of the pipeline’s own API and records each question that the pipeline receives and the answer it returns to the user. These records mark where the run of each question begins and ends, and record the original question and the final answer. Verba queries its Weaviate vector database over gRPC \cite{grpc}, which mitmproxy cannot forward, so we log these calls with a small pass-through gRPC logger in front of Weaviate. The proxy forwards every call unchanged, except when faults are injected (Section~\ref{sec:dataset-injection}), and it sends each call to RAGLog, which classifies it, transforms it into the RAG schema, and assembles the runs (Sections~\ref{sec:classification} and~\ref{sec:transformation}).

For each pipeline and dataset, a launcher sends the questions one at a time through the proxy and checks the completed session: every question has exactly one run with an answer from a generator call, the expected stages are logged, and the logged settings match the harmonized settings. Sessions that failed this check were repeated before any run was labeled.

\subsection{Fault Injection}\label{sec:dataset-injection}
Natural faults are spread unevenly over the stages; in our test runs, embedder faults are rare, and rewriter and reranker faults occur only in the pipelines that have these stages. To evaluate fault localization for every stage with
a known fault, and to study how faults propagate, we built a fault injector that introduces a fault into a single stage of a running pipeline.

\subsubsection{Live injection}
The injector works through the logging proxy. In fault mode, the proxy passes each call to RAGLog before forwarding it. RAGLog classifies the call, maps it to its schema component, and if the call belongs to the stage selected for injection, it changes the input or output of that component. The change is then written back into the request or response in the API's own format, and the proxy forwards the modified call. Since the fault is introduced while the pipeline runs, every later stage receives the faulty input and reacts to it as it would in practice, for example by retrieving other chunks after a faulty rewrite or by answering from a context that lacks the evidence. Faults can only be injected into stages whose calls pass through the proxy; we therefore do not inject retriever faults into RAGFlow and Quivr, whose retrieval runs in-process, or into Verba, whose retriever is logged by the separate gRPC logger. Reranker faults are injected only into RAGFlow and Quivr, since LightRAG's reranker passes on every chunk it ranks.

\subsubsection{Fault types}
\begin{table}[!htbp]
    \centering
    \caption{Fault types injected into the test runs, the pipelines into which they were injected, and the injected
    part of the API call}
    \label{tab:injection}
    \small
    \setlength{\tabcolsep}{4pt}
    \begin{tabular}{@{}>{\raggedright\arraybackslash}p{1.6cm}>{\raggedright\arraybackslash}p{8.0cm}>{\raggedright\arraybackslash}p{2.3cm}>{\raggedright\arraybackslash}p{1.6cm}@{}}
        \toprule
        Stage & Fault types & Pipelines & Target \\
        \midrule
        Rewriter & A wrong year is added; a date or qualifier is dropped; the entity is replaced by a similar one &
            Quivr, Onyx & Response \\
        Embedder & The query is cut to its first half; the key entity is misspelled & All five & Request \\
        Retriever & The evidence chunks are not returned & LightRAG,
            Onyx & Response \\
        Reranker & The evidence is ranked below the cut-off; one of several evidence chunks is ranked below the
            cut-off & RAGFlow, Quivr & Response \\
        Generator & The evidence is placed in the middle of the context among passages about similar entities & All
            five & Request \\
        \bottomrule
    \end{tabular}
\end{table}

Table~\ref{tab:injection} lists the injected fault types. We selected fault types that are reported in other RAG failure studies \cite{barnett2024seven, lauro2025rag, li2024dmqr} and resemble the faults observed in our natural runs. Faults injected into the retriever, reranker, and generator use the question's reference evidence to target the information needed to answer it. For example, a retriever fault removes chunks containing the required evidence from the retrieved set, while a reranker fault moves these chunks just below the cut-off, so they are retrieved but not passed to the generator. For the generator, the fault places the relevant evidence in the middle of the context, surrounded by passages about similar entities retrieved by the same pipeline for other questions \cite{liu2024lost}. The development runs also contain further fault types that we explored and did not use for evaluation because they rarely affected the answer.

\subsubsection{Injected test set}
The injected test set only uses questions on which the pipeline's original run was fault-free. We aimed to include 30 runs for each of the 16 pipeline--stage pairs where faults could be injected, covering all three datasets and different fault types. To construct this set, we injected faults into 634 runs and subsequently labeled them using our judges. However, an injection does not always result in an observable fault, as the pipeline may absorb its effects. For example, a reranker may demote relevant evidence without excluding it from the generator's context, or the generator may ignore distracting passages. The judges, who were unaware of whether or where faults had been injected, confirmed faults at the targeted stage in 345 of the 634 runs. From these labeled runs, we selected a final test set of 480 runs, with 30 runs per pipeline--stage pair, prioritizing runs in which the judges confirmed faults at the injected stage.

\subsection{Labeling}\label{sec:dataset-labels}
We consider a stage faulty when it introduces a fault rather than just passing on an earlier one, regardless of its effect on the final answer. The retriever and reranker are assessed against the evidence needed for the original question, meaning that an earlier pipeline fault may also cause these stages to be labeled faulty; this happens in 4 natural and at most 32 injected test runs. Since labeling every stage of 3,134 runs by hand was not feasible, we used two LLM judges from different providers, GPT-5.1 and Claude Sonnet 5.5, to detect faulty stages. Each judge receives the question, reference answer and evidence, and the full execution trace, with chunks containing the reference evidence marked. For each stage, the judge records factual findings, such as which pieces of information needed to answer the question are present in the retrieved chunks and which statements in the generated answer are not supported by the context. The judge also identifies fault types based on the RAG error taxonomy RAGEC \cite{leung2026classifying}, which we extend with two additional fault types for the rewriter and embedder. 

The stage labels are then determined from these findings using fixed rules. The rewriter is labeled faulty if it adds or removes a constraint from the original question, or changes its intent. The embedder is faulty if the text passed for embedding alters the query, for example, by omitting names or constraints, introducing misspellings, or adding unrelated text. The retriever is faulty if none of the retrieved chunks contains information needed to answer the question. The reranker is faulty if the required information was retrieved but excluded from the chunks passed to the generator. Finally, the generator is faulty if its answer contains unsupported claims, omits explicitly requested information, incorrectly refuses to answer, fails to abstain when the question is unanswerable, or answers a different question. Where both judges agree, their label is final. Where they disagree, a third model, Claude Opus 5.5 \cite{anthropic2026opus55}, adjudicates. Of the 1,500 natural test runs, 1,133 were labeled by agreement and 367 by adjudication. Of the 480 injected test runs, 299 were labeled by agreement and 181 by adjudication.

The judges' labels serve as the ground truth for training and evaluating RAGLog, but they are never an input to it.
RAGLog sees only the trace of a run, without the reference answer and evidence. For an injected run, the injected stage is labeled faulty only if the judges found a fault there; the injection itself records only where a fault was
introduced.

\subsection{Splits, Statistics, and Release}\label{sec:dataset-splits}
Table~\ref{tab:dataset} summarizes the runs of RAGLog Faults. The development runs consist of 750 natural runs, one
for each of the 50 development questions of each dataset on each pipeline, and 250 injected runs on DragonBall
questions. The test runs consist of 1,500 natural runs, one for each of the 100 test questions of each dataset on each pipeline, and the 634 labeled injected runs, of which 480 form the final injected test set.

\begin{table}[!htbp]
    \centering
    \caption{Runs of RAGLog Faults}
    \label{tab:dataset}
    \small
    \setlength{\tabcolsep}{4pt}
  \begin{tabular}{@{}llrrrrrrrr@{}}
        \toprule
        & & \multicolumn{3}{c}{Runs} & \multicolumn{5}{c}{Faulty stage / runs with the stage} \\
        \cmidrule(lr){3-5}\cmidrule(l){6-10}
        & & All & Faulty$^{\dagger}$ & Correct$^{\ast}$ & Rewriter & Embedder & Retriever & Reranker & Generator \\
        \midrule
        Development & Natural & 750 & 264 & 544 & 36/300 & 4/750 & 118/750 & 25/450 & 141/750 \\
                    & Injected & 250 & 188 & 123 & 25/146 & 56/248 & 75/248 & 41/149 & 43/250 \\
        \cmidrule(l){2-10}
        Test        & Natural & 1,500 & 551 & 1,143 & 42/600 & 10/1,500 & 227/1,500 & 62/900 & 329/1,500 \\
                    & Injected & 480 & 333 & 347 & 77/240 & 159/480 & 67/479 & 36/299 & 80/480 \\
        \bottomrule
    \end{tabular}

    \vspace{1mm}
    \footnotesize $^{\dagger}$ Runs with at least one faulty stage. \\ $^{\ast}$ Runs with a correct final answer. 
\end{table}

The natural test runs show how common hidden faults are: 551 of the 1,500 runs (37\%) contain at least one faulty
stage, and 202 of these runs still produce a correct answer. Faults are concentrated in the generator (329 runs) and the retriever (227 runs), and 252 of the 357 incorrect answers (71\%) are caused by a stage before the generator. Embedder faults are rare (10 runs), which is one reason for the injected test set, in which each stage has the same number of injections per pipeline. Among the 634 injected runs that were labeled, almost all rewriter and embedder injections produced a fault at the injected stage, while many retriever, reranker, and generator injections were absorbed. To evaluate localization on known faults while keeping the stages balanced, the final injected test set contains 480 runs (30 runs per pipeline--stage pair), prioritizing runs in which the judges confirmed a fault at the injected stage.

We release RAGLog Faults on Hugging Face.\footnote{\url{https://huggingface.co/datasets/Hira221/raglog-faults}}
It is organized into three subsets, one per source dataset, each distributed under the terms of its source dataset:
CC BY-NC-SA 4.0 for DragonBall, CC BY-SA 4.0 for NQ (whose source is CC BY-SA 3.0), and the MS MARCO terms for non-commercial research. Our annotations are
released under CC BY 4.0, and access requires agreeing to the terms of the source datasets. Each subset contains the
runs with their labels and fault types, the traces, both judges' findings and reasons, and a record of every injection with its fault type.

\section{Experimental Setup}\label{sec:setup}
\subsection{Research Questions}
We answer the following research questions in our evaluation of RAGLog:

\textbf{RQ1: How effective is RAGLog at detecting and localizing natural faults in RAG pipelines?} This question evaluates RAGLog on naturally occurring faults and compares its effectiveness with two general-purpose coding agents, Claude Code and Codex, using the same execution logs. 

\textbf{RQ2: How effective is RAGLog at detecting and localizing injected faults in RAG pipelines?}
    This question evaluates RAGLog on faults injected into individual pipeline stages and compares its effectiveness with the baseline agents. It also examines how injected faults manifest and propagate through the pipeline. 

\textbf{RQ3: How do the key components of RAGLog contribute to its effectiveness?}
    This question examines the contribution of individual components and groups of components to RAGLog's effectiveness.

\textbf{RQ4: How efficient is RAGLog at detecting and localizing faults in RAG pipelines?}
    This question compares RAGLog with the baseline agents in terms of API cost, token consumption, and analysis time.

\subsection{RAGLog Variants}\label{sec:setup-raglog}
We evaluate three variants of RAGLog, each using a different LLM for the LLM-based checks: Claude Haiku 5.5 \cite{anthropic2026haiku55}, GPT-5 Mini \cite{openai2025gpt5mini}, and Gemini 3.1 Flash-Lite \cite{google2026gemini31flashlite}. We denote each variant by its LLM, for example \raglogv{Haiku 5.5}. Claude Haiku 5.5 is RAGLog's default LLM; where we report a single result for RAGLog without a subscript, it refers to \raglogv{Haiku 5.5}. We selected smaller LLMs to keep the cost and latency of fault localization low, making RAGLog suitable for use in production. For each variant, the fault classifiers and their thresholds were trained using the development runs of the RAGLog Faults dataset. The variants were then evaluated on both the natural and injected test sets. All local computations, including the rule-based checks, embedding-based checks, and fault classifiers, were performed on a laptop with an Apple M1 Pro processor and 16\,GB of memory, using Python 3.13 and scikit-learn 1.8.

\subsection{Baselines}\label{sec:baselines}
To the best of our knowledge, no existing approach localizes faults across all stages of a RAG pipeline without reference answers. Existing evaluation frameworks either require reference answers or assess only the
retriever and the generator, while observability platforms record execution traces but leave their analysis to the developer. The diagnostic approaches closest to ours also have similar limitations. RAG-Critic \cite{dong2025rag} provides a critic model that assesses generated responses and attributes faults only to retrieval or generation as a whole, without separating the stages before generation into the rewriter, embedder, retriever, and reranker. Furthermore, it reports faults only when it judges the final answer is incorrect, leaving faults in runs with correct answers undetected. The automatic fault classification of RAGEC \cite{leung2026classifying} requires reference answers. Doctor-RAG \cite{jiao2026doctor} diagnoses agentic RAG runs already known to have failed, and MetaRAG \cite{sok2025metarag} detects hallucinations in the generated answers. A developer without a dedicated tool would therefore most likely hand the logs to a general-purpose coding agent and ask it to find the faulty stages. We use two such agents as baselines: Claude Code \cite{anthropicClaudeCode} with Claude Haiku 5.5, the same model as RAGLog's checks, and Codex \cite{openaiCodex} with GPT-5.5 at medium reasoning effort, a stronger model. Using the same LLM as one of RAGLog's variants helps isolate the contribution of RAGLog's design, while the stronger model tests whether greater reasoning capability can close the performance gap.

Each agent receives the raw proxy logs of a pipeline session containing runs from one dataset, along with a fixed task description. The description defines the five pipeline stages and the criteria for considering each stage faulty, using the same fault definitions as the judges. It also explains the log format and instructs the agent to provide a fault verdict for every stage present in each run. The agents receive no reference answers, access to source documents, or examples from our dataset, and the task description was not adjusted based on test results. Given the size of the logs, the agents write and execute their own scripts to separate the session into individual runs and extract relevant information. Each session is analyzed in a fresh, isolated working directory outside the RAGLog project, ensuring that the agent has access only to the logs and task description. For injected runs, the logs contain the API calls as observed by the pipeline after fault injection, without revealing whether or where a fault was injected. We evaluate the agents' verdicts using the same ground-truth labels and scoring rules as RAGLog.

\subsection{Evaluation Metrics}\label{sec:metrics}
We evaluate fault-localization effectiveness at both the stage and run levels. At the stage level, each stage present in a run is treated as a binary decision: faulty or not faulty. We compare predictions against the ground-truth labels assigned by the judges and report precision, recall, F1 score, and false-positive rate (FPR) for each stage. Precision measures the proportion of flagged stages that are faulty, while recall measures the proportion of faulty stages correctly identified. F1 is the harmonic mean of precision and recall, and FPR measures the proportion of fault-free stage instances incorrectly flagged. To summarize effectiveness across stages, we report macro-F1 and pooled F1. Macro-F1 averages the F1 scores of the stages that have at least one fault in the evaluated runs, giving each stage equal weight. However, faults are unevenly distributed across stages. For example, the natural test runs contain only 10 embedder faults compared with 329 generator faults, making macro-F1 sensitive to a small number of predictions for rarely faulty stages. We therefore use pooled F1 as our primary metric, calculated by aggregating predictions across all stage instances rather than averaging stage-level scores.

To estimate how much pooled F1 depends on the particular test questions, we compute 95\% bootstrap confidence intervals. We draw the test questions with replacement 2,000 times, keep all runs of each sampled question together, and recompute pooled F1 on each sample. To compare RAGLog with a baseline, we compute the interval of the difference between their scores on the same samples; a difference is significant when this interval does not include zero.

We additionally measure run-level detection to assess whether a system identifies faulty runs, regardless of the stage it flags. A run is considered faulty if the judges label at least one stage as faulty, and is flagged if the system predicts a fault in any stage. We report the same four metrics at this level. Along with fault localization, we evaluate the accuracy of RAGLog's API classification and log-to-schema transformation. Classification accuracy is the proportion of calls assigned to the correct RAG stage, including those correctly identified as irrelevant. Since the rule-based classifier may leave some calls undecided, we also report its coverage (the proportion of calls classified) and accuracy on the classified calls. For log-to-schema transformation, accuracy is the proportion of stage calls whose traces contain the expected values, while coverage is the proportion successfully mapped by the transformation rules.

Finally, we evaluate efficiency through token usage, API cost, and execution time. For RAGLog, we record the number of LLM calls, input and output tokens, API cost based on providers' list prices, and analysis time for each run. Since the five LLM calls execute in parallel with local checks, the total analysis time is determined by the slowest operation. For the agent baselines, we record tokens, cost, and wall-clock time for each session, then divide these totals by the number of questions in that session. 

\subsection{Evaluation Procedure}\label{sec:procedure}

\noindent\textit{\textbf{RQ1.}}
We evaluate RAGLog and both agents on the 1,500 natural test runs in our dataset (Section~\ref{sec:dataset-splits}) and compare their effectiveness per stage and per dataset. To assess how well RAGLog generalizes to pipelines it was not trained on, we additionally perform a leave-one-pipeline-out evaluation. For each pipeline, we train the fault classifiers on the development runs of the other four pipelines and apply them to the test runs of the held-out pipeline.

\noindent\textit{\textbf{RQ2.}}
We evaluate RAGLog and both agents on the final injected test set of 480 runs from our dataset (Section~\ref{sec:dataset-splits}) and compare their effectiveness as in RQ1.
To examine how injected faults manifest and propagate through the pipeline, we also analyze all 480 labeled injected runs. For each stage and fault type, we examine how often the judges confirmed a fault at the targeted stage, how often the fault was absorbed, and how often it propagated to later stages or affected the final answer.

\noindent\textit{\textbf{RQ3.}}
We evaluate the three components of RAGLog, corresponding to its three steps. Since fault localization depends on API call classification and log-to-schema transformation, these components cannot be removed and are instead evaluated directly on familiar and unfamiliar APIs. For fault localization, we assess the contribution of its individual parts through an ablation study.

\textit{API call classification.} We use two sets of labeled API calls. The first, which we call \textit{familiar APIs}, contains all recorded calls from the five evaluation pipelines (RAGFlow, LightRAG, Quivr, Onyx, and Verba). The second, \textit{unfamiliar APIs}, contains calls to providers and endpoints not used by any of these pipelines, covering embedding services, vector databases, rerankers, and LLM APIs, as well as calls unrelated to RAG stages, such as ingestion, setup, title generation, and routing. These calls follow the providers' documented API formats or were collected from live API requests to reflect realistic traffic. Each call is labeled with its corresponding stage or as irrelevant. For both sets, we evaluate three classification methods: rules alone, the LLM alone, and the hybrid approach, where the LLM handles calls not recognized by the rules. We measure classification accuracy for each method, the proportion of calls covered by the rules, and the number of LLM calls required by the combined approach.

\textit{Log-to-schema transformation.} We use the same two sets of API calls. For both sets, we generate mappings for individual stage calls and compare the extracted values, including queries, retrieved chunks, rankings, context, and answers, against manually established reference values. We evaluate the transformation accuracy of three methods: rules alone, the LLM alone, and the hybrid approach, where the LLM generates mappings for calls not covered by the rules. We evaluate the LLM-based components of both API call classification and log-to-schema transformation using Claude Haiku 5.5 and GPT-4.1 mini.

\textit{Fault localization.} We assess the contributions of the stage checks and ensemble decision through an ablation study, modifying one component at a time.

\begin{itemize}
    \item \textit{Stage checks.} We remove each group of checks in Table~\ref{tab:checks} (rule-based, embedding-based, and LLM-based), and then each of the five LLM-based checks individually.
    \item \textit{Ensemble decision.} We examine four design choices: replacing the ensemble with each of its three model types individually, varying the model weights, restricting each stage classifier to measurements from its own stage to assess the benefit of cross-stage information, and removing the false-alarm cap, the 10\% limit on the false-positive rate used to select decision thresholds.
\end{itemize}
For each configuration, we retrain the fault classifiers on the development runs and evaluate their effectiveness on the test runs.

\noindent\textit{\textbf{RQ4.}}
We compare RAGLog and both agents in terms of API cost, token usage, and analysis time per question across the natural and injected test runs.

\section{Results}\label{sec:results}

\subsection{RQ1: How effective is RAGLog at detecting and localizing natural faults in RAG pipelines?}\label{sec:rq1}

\begin{table}[!htbp]
    \centering
    \caption{Effectiveness on natural runs}
    \label{tab:rq1}
    \small
    \setlength{\tabcolsep}{3pt}
    \begin{tabular}{@{}llrrrrrrrrrrrrrrr@{}}
        \toprule
        & & \multicolumn{3}{c}{Rewriter} & \multicolumn{3}{c}{Embedder} & \multicolumn{3}{c}{Retriever} & \multicolumn{3}{c}{Reranker} & \multicolumn{3}{c}{Generator} \\
        \cmidrule(lr){3-5}\cmidrule(lr){6-8}\cmidrule(lr){9-11}\cmidrule(lr){12-14}\cmidrule(lr){15-17}
        Dataset & System & P & R & F1 & P & R & F1 & P & R & F1 & P & R & F1 & P & R & F1 \\
        \midrule
        DragonBall & Claude Code\textsubscript{Haiku 5.5} & \textbf{1.00} & 0.07 & 0.13 & 0.00 & 0.00 & 0.00 & \textbf{0.91} & 0.84 & \textbf{0.87} & 0.83 & 0.62 & 0.71 & \textbf{0.87} & 0.49 & 0.63 \\
         & Codex\textsubscript{GPT-5.5} & 0.05 & 0.07 & 0.06 & 0.00 & 0.00 & 0.00 & 0.88 & 0.84 & 0.86 & 0.82 & 0.57 & 0.68 & \textbf{0.87} & 0.41 & 0.56 \\
         & RAGLog\textsubscript{Gemini 3.1 Flash-Lite} & 0.57 & \textbf{0.93} & 0.70 & \textbf{0.25} & \textbf{0.11} & \textbf{0.15} & 0.75 & 0.91 & 0.83 & \textbf{0.88} & 0.55 & 0.68 & 0.66 & 0.66 & 0.66 \\
         & RAGLog\textsubscript{GPT-5 Mini} & 0.54 & \textbf{0.93} & 0.68 & \textbf{0.25} & \textbf{0.11} & \textbf{0.15} & 0.77 & 0.90 & 0.83 & 0.66 & \textbf{0.78} & 0.71 & 0.77 & 0.68 & 0.72 \\
         & RAGLog\textsubscript{Haiku 5.5} & 0.62 & \textbf{0.93} & \textbf{0.74} & 0.00 & 0.00 & 0.00 & 0.77 & \textbf{0.92} & 0.84 & 0.77 & 0.75 & \textbf{0.76} & 0.68 & \textbf{0.81} & \textbf{0.74} \\
        \midrule
        NQ & Claude Code\textsubscript{Haiku 5.5} & \textbf{1.00} & 0.29 & 0.45 & 0.00 & 0.00 & 0.00 & \textbf{0.93} & 0.37 & 0.53 & \textbf{0.86} & 0.40 & 0.55 & 0.63 & 0.41 & 0.50 \\
         & Codex\textsubscript{GPT-5.5} & 0.00 & 0.00 & 0.00 & 0.00 & 0.00 & 0.00 & 0.61 & 0.57 & 0.59 & 0.75 & 0.20 & 0.32 & 0.73 & 0.19 & 0.30 \\
         & RAGLog\textsubscript{Gemini 3.1 Flash-Lite} & 0.27 & 0.35 & 0.31 & 0.00 & 0.00 & 0.00 & 0.66 & 0.71 & 0.68 & 0.69 & 0.60 & 0.64 & 0.68 & 0.60 & 0.63 \\
         & RAGLog\textsubscript{GPT-5 Mini} & 0.42 & \textbf{0.65} & \textbf{0.51} & 0.00 & 0.00 & 0.00 & 0.58 & 0.63 & 0.60 & 0.47 & 0.60 & 0.53 & \textbf{0.77} & 0.54 & 0.63 \\
         & RAGLog\textsubscript{Haiku 5.5} & 0.37 & 0.59 & 0.45 & 0.00 & 0.00 & 0.00 & 0.68 & \textbf{0.80} & \textbf{0.74} & 0.55 & \textbf{0.80} & \textbf{0.65} & 0.75 & \textbf{0.71} & \textbf{0.73} \\
        \midrule
        MS MARCO & Claude Code\textsubscript{Haiku 5.5} & \textbf{0.60} & 0.27 & 0.38 & -- & -- & -- & \textbf{0.64} & 0.35 & 0.45 & 0.00 & 0.00 & 0.00 & 0.64 & 0.21 & 0.31 \\
         & Codex\textsubscript{GPT-5.5} & 0.00 & 0.00 & 0.00 & -- & -- & -- & 0.55 & 0.35 & 0.42 & 0.00 & 0.00 & 0.00 & \textbf{0.83} & 0.16 & 0.27 \\
         & RAGLog\textsubscript{Gemini 3.1 Flash-Lite} & 0.25 & 0.45 & 0.32 & -- & -- & -- & 0.57 & 0.75 & 0.65 & \textbf{0.75} & \textbf{0.86} & \textbf{0.80} & 0.67 & 0.57 & 0.62 \\
         & RAGLog\textsubscript{GPT-5 Mini} & 0.22 & 0.36 & 0.28 & -- & -- & -- & 0.58 & 0.75 & 0.66 & 0.40 & \textbf{0.86} & 0.55 & 0.81 & 0.59 & 0.68 \\
         & RAGLog\textsubscript{Haiku 5.5} & 0.31 & \textbf{0.82} & \textbf{0.45} & -- & -- & -- & 0.61 & \textbf{0.77} & \textbf{0.68} & 0.55 & \textbf{0.86} & 0.67 & 0.75 & \textbf{0.73} & \textbf{0.74} \\
        \midrule
        All & Claude Code\textsubscript{Haiku 5.5} & \textbf{0.82} & 0.21 & 0.34 & 0.00 & 0.00 & 0.00 & \textbf{0.87} & 0.65 & 0.75 & \textbf{0.84} & 0.50 & 0.63 & 0.72 & 0.36 & 0.48 \\
         & Codex\textsubscript{GPT-5.5} & 0.05 & 0.02 & 0.03 & 0.00 & 0.00 & 0.00 & 0.78 & 0.69 & 0.73 & 0.81 & 0.42 & 0.55 & \textbf{0.82} & 0.25 & 0.39 \\
         & RAGLog\textsubscript{Gemini 3.1 Flash-Lite} & 0.37 & 0.57 & 0.45 & \textbf{0.25} & \textbf{0.10} & \textbf{0.14} & 0.70 & 0.85 & 0.76 & 0.80 & 0.60 & 0.69 & 0.67 & 0.61 & 0.64 \\
         & RAGLog\textsubscript{GPT-5 Mini} & 0.41 & 0.67 & 0.51 & \textbf{0.25} & \textbf{0.10} & \textbf{0.14} & 0.70 & 0.82 & 0.76 & 0.57 & 0.74 & 0.64 & 0.78 & 0.60 & 0.68 \\
         & RAGLog\textsubscript{Haiku 5.5} & 0.42 & \textbf{0.76} & \textbf{0.54} & 0.00 & 0.00 & 0.00 & 0.72 & \textbf{0.87} & \textbf{0.79} & 0.67 & \textbf{0.77} & \textbf{0.72} & 0.72 & \textbf{0.75} & \textbf{0.74} \\
        \bottomrule
    \end{tabular}
    
    \vspace{1mm}
    \footnotesize P: precision; R: recall; F1: F1 score. \\ Based on 1,500 natural test runs; stage-level sample sizes vary because not all pipelines contain every stage.\\ --: no fault of this stage in the runs.
\end{table}

\begin{table}[!htbp]
    \centering
    \caption{Stage-level false-positive rates (\%), macro-F1, and pooled F1 on natural runs}
    \label{tab:rq1-fpr}
    \small
    \setlength{\tabcolsep}{3pt}
    \begin{tabular}{@{}llrrrrrrrr@{}}
        \toprule
        & & \multicolumn{6}{c}{False-positive rate (\%)} & \multicolumn{2}{c}{F1} \\
        \cmidrule(lr){3-8}\cmidrule(lr){9-10}
        Dataset & System & Rewriter & Embedder & Retriever & Reranker & Generator & Pooled & Macro & Pooled \\
        \midrule
        DragonBall & Claude Code\textsubscript{Haiku 5.5} & \textbf{0.0} & \textbf{0.0} & \textbf{3.1} & 1.9 & 2.1 & \textbf{1.4} & 0.47 & 0.74 \\
         & Codex\textsubscript{GPT-5.5} & 9.7 & \textbf{0.0} & 4.4 & 1.9 & \textbf{1.8} & 2.7 & 0.43 & 0.68 \\
         & RAGLog\textsubscript{Gemini 3.1 Flash-Lite} & 5.4 & 0.6 & 11.7 & \textbf{1.2} & 9.8 & 5.7 & 0.60 & 0.73 \\
         & RAGLog\textsubscript{GPT-5 Mini} & 5.9 & 0.6 & 10.3 & 6.2 & 5.7 & 5.3 & \textbf{0.62} & 0.76 \\
         & RAGLog\textsubscript{Haiku 5.5} & 4.3 & 0.8 & 10.6 & 3.5 & 10.8 & 6.0 & \textbf{0.62} & \textbf{0.77} \\
        \midrule
        NQ & Claude Code\textsubscript{Haiku 5.5} & \textbf{0.0} & \textbf{0.0} & \textbf{0.2} & \textbf{0.4} & 6.1 & 1.4 & 0.41 & 0.50 \\
         & Codex\textsubscript{GPT-5.5} & \textbf{0.0} & \textbf{0.0} & 2.8 & \textbf{0.4} & \textbf{1.8} & \textbf{1.2} & 0.24 & 0.37 \\
         & RAGLog\textsubscript{Gemini 3.1 Flash-Lite} & 8.7 & \textbf{0.0} & 2.8 & 1.4 & 7.1 & 3.3 & 0.45 & 0.61 \\
         & RAGLog\textsubscript{GPT-5 Mini} & 8.2 & \textbf{0.0} & 3.4 & 3.5 & 4.1 & 3.1 & 0.45 & 0.60 \\
         & RAGLog\textsubscript{Haiku 5.5} & 9.3 & \textbf{0.0} & 2.8 & 3.5 & 5.9 & 3.5 & \textbf{0.51} & \textbf{0.69} \\
        \midrule
        MS MARCO & Claude Code\textsubscript{Haiku 5.5} & 1.1 & \textbf{0.0} & \textbf{2.2} & \textbf{0.0} & 3.7 & 1.4 & 0.28 & 0.35 \\
         & Codex\textsubscript{GPT-5.5} & \textbf{0.0} & \textbf{0.0} & 3.3 & \textbf{0.0} & \textbf{1.1} & \textbf{1.1} & 0.17 & 0.30 \\
         & RAGLog\textsubscript{Gemini 3.1 Flash-Lite} & 7.9 & \textbf{0.0} & 6.5 & 0.7 & 9.1 & 4.4 & 0.60 & 0.61 \\
         & RAGLog\textsubscript{GPT-5 Mini} & 7.4 & \textbf{0.0} & 6.2 & 3.1 & 4.5 & 3.8 & 0.54 & 0.63 \\
         & RAGLog\textsubscript{Haiku 5.5} & 10.6 & 0.2 & 5.8 & 1.7 & 7.7 & 4.5 & \textbf{0.63} & \textbf{0.69} \\
        \midrule
        All & Claude Code\textsubscript{Haiku 5.5} & \textbf{0.4} & \textbf{0.0} & \textbf{1.7} & \textbf{0.7} & 4.0 & \textbf{1.4} & 0.44 & 0.58 \\
         & Codex\textsubscript{GPT-5.5} & 3.2 & \textbf{0.0} & 3.5 & \textbf{0.7} & \textbf{1.6} & 1.6 & 0.34 & 0.52 \\
         & RAGLog\textsubscript{Gemini 3.1 Flash-Lite} & 7.3 & 0.2 & 6.6 & 1.1 & 8.6 & 4.5 & 0.54 & 0.67 \\
         & RAGLog\textsubscript{GPT-5 Mini} & 7.2 & 0.2 & 6.4 & 4.2 & 4.8 & 4.0 & 0.55 & 0.69 \\
         & RAGLog\textsubscript{Haiku 5.5} & 8.1 & 0.3 & 6.0 & 2.9 & 8.1 & 4.6 & \textbf{0.56} & \textbf{0.73} \\
        \bottomrule
    \end{tabular}
\end{table}

Tables~\ref{tab:rq1} to~\ref{tab:rq1-run} show the effectiveness of the three variants of RAGLog and the two agent baselines on a test set of 1,500 natural runs. We report precision, recall, and F1 score for each stage (Table~\ref{tab:rq1}), false-positive rates, macro-F1, and pooled F1 (Table~\ref{tab:rq1-fpr}), and run-level detection (Table~\ref{tab:rq1-run}). Overall, \raglogv{Haiku 5.5} achieves a pooled F1 score of 0.73, compared to 0.58 for Claude Code and 0.52 for Codex, and a macro-F1 score of 0.56, compared to 0.44 and 0.34, respectively. RAGLog also outperforms the agents with its other variants, achieving pooled F1 scores of 0.69 for \raglogv{GPT-5 Mini} and 0.67 for \raglogv{Gemini 3.1 Flash-Lite}. These improvements are statistically significant, with 95\% bootstrap confidence intervals for the pooled F1 difference of [0.10, 0.19] against Claude Code and [0.16, 0.25] against Codex for \raglogv{Haiku 5.5}. The other two variants also significantly outperform both agents.  Since \raglogv{Haiku 5.5} and Claude Code use the same LLM, the difference between their performance cannot be attributed to the underlying model alone, suggesting that RAGLog’s design contributes to its improved performance.

\begin{figure}[!htbp]
    \centering
    \includegraphics[width=\linewidth]{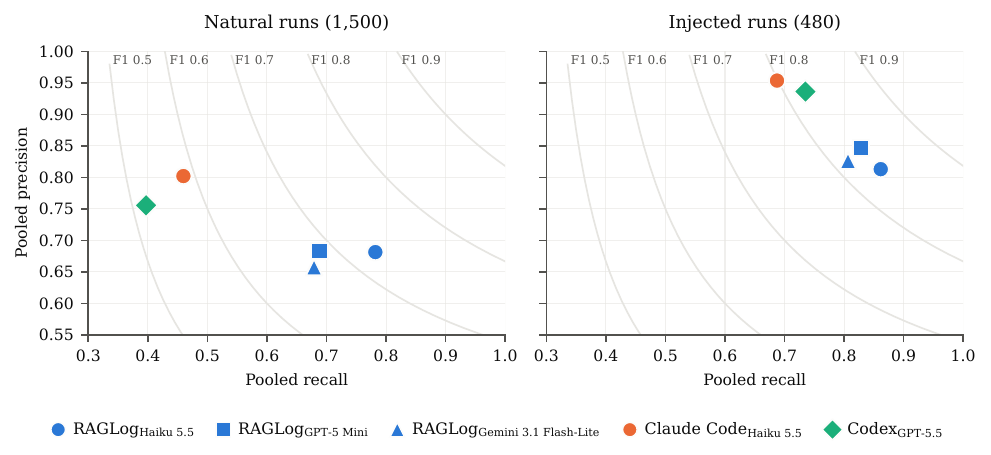}
    \caption{Pooled precision and recall on natural and injected runs; curves show equal F1 scores.}
    \label{fig:pooled-pr}
\end{figure}

The main difference between RAGLog and the agents is recall. The agents flag relatively few stages, but most of their predictions are correct. Claude Code achieves a pooled precision of 0.80 and Codex 0.76, with false-positive rates of 1.4\% and 1.6\%, respectively. However, their pooled recall is only 0.46 and 0.40 (Figure~\ref{fig:pooled-pr}). In contrast, \raglogv{Haiku 5.5} detects 78\% of faulty stages, with a precision of 0.68 and a false-positive rate of 4.6\%.  RAGLog therefore trades a higher false-alarm rate for substantially higher recall. 

\subsubsection{Stage-level effectiveness}
As shown in Table \ref{tab:rq1}, the largest differences occur at the generator and rewriter stages. For the generator, the most frequently faulty stage (329 runs), \raglogv{Haiku 5.5} achieves an F1 score of 0.74, compared to 0.48 for Claude Code and 0.39 for Codex. The agents' generator recall is relatively low, at 0.36 and 0.25, respectively. Generator faults, such as introducing unsupported facts or omitting information requested by the question, require comparing the generated answer against its context. RAGLog performs these checks explicitly for every run. For the rewriter, \raglogv{Haiku 5.5} achieves an F1 score of 0.54, compared to 0.34 for Claude Code and 0.03 for Codex, which rarely identifies faulty rewrites. Retriever performance is relatively similar across systems, with F1 scores ranging from 0.73 to 0.79. For the reranker, RAGLog achieves F1 scores of 0.64--0.72 across its variants, compared to 0.63 for Claude Code and 0.55 for Codex. Natural embedder faults are rare (10 runs) and difficult for all systems. Neither agent nor \raglogv{Haiku 5.5} detects any, while \raglogv{GPT-5 Mini} and \raglogv{Gemini 3.1 Flash-Lite} each detect one (F1 score of 0.14).

\subsubsection{Effectiveness across datasets}
RAGLog's advantage varies across datasets (Table~\ref{tab:rq1-fpr}). On DragonBall, which contains generated questions and documents, the gap is smallest: \raglogv{Haiku 5.5} achieves a pooled F1 score of 0.77, compared to 0.74 for Claude Code, which performs better at retrieval. On NQ and MS MARCO, which contain real-world questions and documents, Claude Code's pooled F1 drops to 0.50 and 0.35, respectively, while \raglogv{Haiku 5.5} achieves 0.69 on both. One possible explanation is that RAGLog applies the same targeted stage checks across different document types, whereas the agents must independently extract and interpret relevant information from the logs.

\begin{table}[!htbp]
    \centering
    \caption{Run-level detection on natural runs}
    \label{tab:rq1-run}
    \small
    \setlength{\tabcolsep}{3pt}
    \begin{tabular}{@{}lrrrrrrrrrrrrrrrr@{}}
        \toprule
        & \multicolumn{4}{c}{DragonBall} & \multicolumn{4}{c}{NQ} & \multicolumn{4}{c}{MS MARCO} & \multicolumn{4}{c}{All} \\
        \cmidrule(lr){2-5}\cmidrule(lr){6-9}\cmidrule(lr){10-13}\cmidrule(lr){14-17}
        System & P & R & F1 & FPR & P & R & F1 & FPR & P & R & F1 & FPR & P & R & F1 & FPR \\
        \midrule
        Claude Code\textsubscript{Haiku 5.5} & \textbf{0.93} & 0.71 & 0.80 & \textbf{5.2} & 0.74 & 0.45 & 0.56 & 6.4 & 0.78 & 0.33 & 0.47 & 4.4 & 0.85 & 0.53 & 0.66 & 5.4 \\
        Codex\textsubscript{GPT-5.5} & 0.88 & 0.70 & 0.78 & 10.0 & \textbf{0.88} & 0.30 & 0.45 & \textbf{1.7} & \textbf{0.86} & 0.27 & 0.41 & \textbf{2.1} & \textbf{0.87} & 0.47 & 0.61 & \textbf{4.0} \\
        RAGLog\textsubscript{Gemini 3.1 Flash-Lite} & 0.84 & 0.87 & 0.85 & 16.5 & 0.70 & 0.67 & 0.68 & 11.1 & 0.70 & 0.70 & 0.70 & 14.1 & 0.77 & 0.77 & 0.77 & 13.6 \\
        RAGLog\textsubscript{GPT-5 Mini} & 0.85 & 0.87 & 0.86 & 15.7 & 0.76 & 0.64 & 0.69 & 7.8 & 0.79 & 0.72 & 0.75 & 9.1 & 0.81 & 0.77 & 0.79 & 10.3 \\
        RAGLog\textsubscript{Haiku 5.5} & 0.83 & \textbf{0.90} & \textbf{0.87} & 18.1 & 0.74 & \textbf{0.76} & \textbf{0.75} & 10.6 & 0.74 & \textbf{0.81} & \textbf{0.77} & 13.5 & 0.78 & \textbf{0.84} & \textbf{0.81} & 13.6 \\
        \bottomrule
    \end{tabular}
    
    \vspace{1mm}
    \footnotesize P: precision; R: recall; F1: F1 score; FPR: false-positive rate. \\Based on 1,500 natural test runs; a run is faulty if at least one stage is faulty.
\end{table}

\subsubsection{Run-level detection}
Table \ref{tab:rq1-run} shows that at run level, \raglogv{Haiku 5.5} achieves an F1 score of 0.81, compared to 0.66 for Claude Code and 0.61 for Codex. It identifies at least one faulty stage in 84\% of runs containing a fault, compared to 53\% for Claude Code and 47\% for Codex. This indicates that RAGLog is more effective at detecting whether a run contains a fault than at identifying every faulty stage correctly. However, its higher run-level recall comes with a false-positive rate of 13.6\%, compared to 5.4\% and 4.0\% for the agents.

\subsubsection{Effect of the LLM}
Among the three LLMs, Claude Haiku 5.5 achieves the highest stage-level pooled F1 (0.73) and run-level F1 (0.81), respectively. This is followed by GPT-5 Mini at 0.69 stage-level and 0.79 run-level and Gemini 3.1 Flash-Lite at 0.67 stage-level and 0.77 run-level. Their differences are particularly visible at the generator stage. \raglogv{GPT-5 Mini} achieves the highest precision (0.78), but lower recall (0.60), while \raglogv{Haiku 5.5} achieves a more balanced precision of 0.72 and recall of 0.75. 

\subsubsection{Generalization to unseen pipelines}
To evaluate generalization to unseen pipelines, we train RAGLog's fault classifiers on development runs from four pipelines and evaluate them on the test runs of the fifth. On pipelines that were not seen during training, the pooled F1 score of \raglogv{Haiku 5.5} decreases from 0.73 to 0.67 on natural runs, while its false-positive rate remains similar (4.6\% and 4.3\%). However, its pooled F1 remains higher than those of Claude Code (0.58) and Codex (0.52). The other variants also experience decreases, reaching 0.61 for \raglogv{GPT-5 Mini} and 0.60 for \raglogv{Gemini 3.1 Flash-Lite}, which is still higher than both agents. These results suggest that RAGLog can generalize to pipelines not represented in its training data, although its effectiveness decreases. RAGLog can be trained on data from more pipelines in the future.

\begin{answerbox}{Answer to RQ1:}
On natural runs, \raglogv{Haiku 5.5}
localizes faults more effectively than both agent baselines, with a pooled F1 score of 0.73 compared to 0.58 for Claude Code and
0.52 for Codex. Its advantage comes primarily from substantially higher recall, particularly at the generator and rewriter stages and on the real-world questions of NQ and MS MARCO, while the
agents are competitive on DragonBall. Among the three LLMs, Claude Haiku 5.5 performs best overall, with the highest pooled F1 score. All three variants also perform better than both agents on pipelines excluded from their training data (0.60--0.67).
\end{answerbox}

\subsection{RQ2: How effective is RAGLog at detecting and localizing injected faults in RAG pipelines?}\label{sec:rq2}

\begin{table}[!htbp]
    \centering
    \caption{Effectiveness on injected runs}
    \label{tab:rq2}
    \small
    \setlength{\tabcolsep}{3pt}
    \begin{tabular}{@{}llrrrrrrrrrrrrrrr@{}}
        \toprule
        & & \multicolumn{3}{c}{Rewriter} & \multicolumn{3}{c}{Embedder} & \multicolumn{3}{c}{Retriever} & \multicolumn{3}{c}{Reranker} & \multicolumn{3}{c}{Generator} \\
        \cmidrule(lr){3-5}\cmidrule(lr){6-8}\cmidrule(lr){9-11}\cmidrule(lr){12-14}\cmidrule(lr){15-17}
        Dataset & System & P & R & F1 & P & R & F1 & P & R & F1 & P & R & F1 & P & R & F1 \\
        \midrule
        DragonBall & Claude Code\textsubscript{Haiku 5.5} & \textbf{1.00} & 0.77 & 0.87 & \textbf{1.00} & 0.88 & 0.94 & \textbf{0.88} & 0.30 & 0.45 & 0.95 & 0.86 & \textbf{0.90} & 0.81 & 0.39 & 0.53 \\
         & Codex\textsubscript{GPT-5.5} & \textbf{1.00} & 0.77 & 0.87 & \textbf{1.00} & \textbf{0.92} & \textbf{0.96} & \textbf{0.88} & \textbf{0.65} & \textbf{0.75} & \textbf{1.00} & 0.71 & 0.83 & \textbf{0.88} & 0.45 & 0.60 \\
         & RAGLog\textsubscript{Gemini 3.1 Flash-Lite} & \textbf{1.00} & 0.96 & \textbf{0.98} & \textbf{1.00} & 0.87 & 0.93 & 0.44 & 0.52 & 0.48 & 0.90 & 0.90 & \textbf{0.90} & 0.64 & \textbf{0.76} & 0.69 \\
         & RAGLog\textsubscript{GPT-5 Mini} & 0.93 & \textbf{1.00} & 0.96 & \textbf{1.00} & 0.90 & 0.95 & 0.52 & 0.52 & 0.52 & 0.75 & 0.86 & 0.80 & 0.72 & 0.70 & 0.71 \\
         & RAGLog\textsubscript{Haiku 5.5} & 0.96 & \textbf{1.00} & \textbf{0.98} & 0.94 & 0.90 & 0.92 & 0.58 & \textbf{0.65} & 0.61 & 0.77 & \textbf{0.95} & 0.85 & 0.69 & \textbf{0.76} & \textbf{0.72} \\
        \midrule
        NQ & Claude Code\textsubscript{Haiku 5.5} & \textbf{1.00} & 0.80 & \textbf{0.89} & \textbf{1.00} & 0.82 & 0.90 & \textbf{1.00} & 0.44 & 0.61 & 0.67 & 0.67 & 0.67 & \textbf{0.83} & 0.53 & 0.65 \\
         & Codex\textsubscript{GPT-5.5} & 0.91 & 0.84 & 0.88 & \textbf{1.00} & 0.91 & 0.95 & \textbf{1.00} & 0.62 & 0.77 & \textbf{1.00} & 0.33 & 0.50 & 0.75 & 0.32 & 0.44 \\
         & RAGLog\textsubscript{Gemini 3.1 Flash-Lite} & 0.88 & \textbf{0.88} & 0.88 & \textbf{1.00} & 0.91 & 0.95 & 0.91 & 0.62 & 0.74 & 0.75 & \textbf{1.00} & \textbf{0.86} & 0.48 & 0.58 & 0.52 \\
         & RAGLog\textsubscript{GPT-5 Mini} & 0.78 & 0.84 & 0.81 & \textbf{1.00} & \textbf{0.93} & \textbf{0.96} & \textbf{1.00} & \textbf{0.81} & \textbf{0.90} & 0.60 & \textbf{1.00} & 0.75 & 0.67 & 0.63 & 0.65 \\
         & RAGLog\textsubscript{Haiku 5.5} & 0.80 & 0.80 & 0.80 & \textbf{1.00} & 0.87 & 0.93 & 0.86 & 0.75 & 0.80 & 0.38 & \textbf{1.00} & 0.55 & 0.64 & \textbf{0.74} & \textbf{0.68} \\
        \midrule
        MS MARCO & Claude Code\textsubscript{Haiku 5.5} & \textbf{1.00} & 0.81 & 0.89 & \textbf{1.00} & \textbf{0.98} & \textbf{0.99} & 0.82 & 0.50 & 0.62 & 0.80 & 0.33 & 0.47 & \textbf{0.83} & 0.36 & 0.50 \\
         & Codex\textsubscript{GPT-5.5} & \textbf{1.00} & \textbf{0.85} & \textbf{0.92} & \textbf{1.00} & \textbf{0.98} & \textbf{0.99} & 0.79 & 0.68 & 0.73 & 0.75 & 0.50 & 0.60 & 0.60 & 0.32 & 0.42 \\
         & RAGLog\textsubscript{Gemini 3.1 Flash-Lite} & 0.91 & 0.77 & 0.83 & \textbf{1.00} & 0.96 & 0.98 & 0.76 & 0.89 & 0.82 & \textbf{0.86} & 0.50 & 0.63 & 0.56 & 0.54 & 0.55 \\
         & RAGLog\textsubscript{GPT-5 Mini} & 0.91 & 0.77 & 0.83 & \textbf{1.00} & \textbf{0.98} & \textbf{0.99} & \textbf{0.87} & 0.96 & \textbf{0.92} & 0.67 & 0.67 & 0.67 & 0.58 & 0.54 & 0.56 \\
         & RAGLog\textsubscript{Haiku 5.5} & 0.81 & 0.81 & 0.81 & \textbf{1.00} & 0.96 & 0.98 & 0.85 & \textbf{1.00} & \textbf{0.92} & 0.69 & \textbf{0.92} & \textbf{0.79} & 0.57 & \textbf{0.75} & \textbf{0.65} \\
        \midrule
        All & Claude Code\textsubscript{Haiku 5.5} & \textbf{1.00} & 0.79 & 0.88 & \textbf{1.00} & 0.89 & 0.94 & \textbf{0.88} & 0.42 & 0.57 & 0.89 & 0.67 & 0.76 & \textbf{0.82} & 0.41 & 0.55 \\
         & Codex\textsubscript{GPT-5.5} & 0.97 & 0.82 & 0.89 & \textbf{1.00} & \textbf{0.94} & \textbf{0.97} & 0.86 & 0.66 & 0.75 & \textbf{0.92} & 0.61 & 0.73 & 0.75 & 0.38 & 0.50 \\
         & RAGLog\textsubscript{Gemini 3.1 Flash-Lite} & 0.93 & \textbf{0.87} & \textbf{0.90} & \textbf{1.00} & 0.91 & 0.95 & 0.66 & 0.70 & 0.68 & 0.88 & 0.78 & \textbf{0.82} & 0.57 & 0.64 & 0.60 \\
         & RAGLog\textsubscript{GPT-5 Mini} & 0.87 & \textbf{0.87} & 0.87 & \textbf{1.00} & \textbf{0.94} & \textbf{0.97} & 0.78 & 0.78 & 0.78 & 0.71 & 0.81 & 0.75 & 0.66 & 0.62 & 0.64 \\
         & RAGLog\textsubscript{Haiku 5.5} & 0.86 & \textbf{0.87} & 0.86 & 0.98 & 0.91 & 0.94 & 0.75 & \textbf{0.82} & \textbf{0.79} & 0.68 & \textbf{0.94} & 0.79 & 0.63 & \textbf{0.75} & \textbf{0.69} \\
        \bottomrule
    \end{tabular}
    
    \vspace{1mm}
    \footnotesize P: precision; R: recall; F1: F1 score. \\Based on 480 injected test runs; stage-level sample sizes vary by pipeline and injection type. 
\end{table}

\begin{table}[!htbp]
    \centering
    \caption{Stage-level false-positive rates (\%), macro-F1, and pooled F1 on injected runs}
    \label{tab:rq2-fpr}
    \small
    \setlength{\tabcolsep}{3pt}
    \begin{tabular}{@{}llrrrrrrrrr@{}}
        \toprule
        & & \multicolumn{6}{c}{False-positive rate (\%)} & \multicolumn{2}{c}{F1} &  \\
        \cmidrule(lr){3-8}\cmidrule(lr){9-10}
        Dataset & System & Rewriter & Embedder & Retriever & Reranker & Generator & Pooled & Macro & Pooled & Conf (\%) \\
        \midrule
        DragonBall & Claude Code\textsubscript{Haiku 5.5} & \textbf{0.0} & \textbf{0.0} & \textbf{0.8} & 1.4 & 2.5 & 1.1 & 0.74 & 0.79 & 79 \\
         & Codex\textsubscript{GPT-5.5} & \textbf{0.0} & \textbf{0.0} & 1.5 & \textbf{0.0} & \textbf{1.7} & \textbf{0.8} & 0.80 & \textbf{0.83} & \textbf{88} \\
         & RAGLog\textsubscript{Gemini 3.1 Flash-Lite} & \textbf{0.0} & \textbf{0.0} & 11.5 & 2.9 & 11.8 & 6.5 & 0.80 & 0.81 & 84 \\
         & RAGLog\textsubscript{GPT-5 Mini} & 3.5 & \textbf{0.0} & 8.5 & 8.7 & 7.6 & 5.9 & 0.79 & 0.82 & 85 \\
         & RAGLog\textsubscript{Haiku 5.5} & 1.8 & 3.0 & 8.5 & 8.7 & 9.2 & 6.7 & \textbf{0.82} & \textbf{0.83} & \textbf{88} \\
        \midrule
        NQ & Claude Code\textsubscript{Haiku 5.5} & \textbf{0.0} & \textbf{0.0} & \textbf{0.0} & 1.0 & \textbf{1.5} & \textbf{0.6} & 0.74 & 0.82 & 83 \\
         & Codex\textsubscript{GPT-5.5} & 4.1 & \textbf{0.0} & \textbf{0.0} & \textbf{0.0} & \textbf{1.5} & 0.8 & 0.71 & 0.84 & 89 \\
         & RAGLog\textsubscript{Gemini 3.1 Flash-Lite} & 6.1 & \textbf{0.0} & 0.7 & 1.0 & 9.0 & 3.3 & 0.79 & 0.83 & 87 \\
         & RAGLog\textsubscript{GPT-5 Mini} & 12.2 & \textbf{0.0} & \textbf{0.0} & 2.1 & 4.5 & 2.7 & \textbf{0.81} & \textbf{0.86} & \textbf{93} \\
         & RAGLog\textsubscript{Haiku 5.5} & 10.2 & \textbf{0.0} & 1.4 & 5.2 & 6.0 & 3.8 & 0.75 & 0.83 & 89 \\
        \midrule
        MS MARCO & Claude Code\textsubscript{Haiku 5.5} & \textbf{0.0} & \textbf{0.0} & \textbf{2.1} & \textbf{1.0} & \textbf{1.4} & \textbf{1.1} & 0.70 & 0.79 & 79 \\
         & Codex\textsubscript{GPT-5.5} & \textbf{0.0} & \textbf{0.0} & 3.5 & 2.0 & 4.2 & 2.3 & 0.73 & 0.80 & 85 \\
         & RAGLog\textsubscript{Gemini 3.1 Flash-Lite} & 3.5 & \textbf{0.0} & 5.6 & \textbf{1.0} & 8.5 & 4.1 & 0.76 & 0.81 & 86 \\
         & RAGLog\textsubscript{GPT-5 Mini} & 3.5 & \textbf{0.0} & 2.8 & 4.1 & 7.7 & 3.8 & 0.79 & 0.84 & 90 \\
         & RAGLog\textsubscript{Haiku 5.5} & 8.8 & \textbf{0.0} & 3.5 & 5.1 & 11.3 & 5.6 & \textbf{0.83} & \textbf{0.85} & \textbf{94} \\
        \midrule
        All & Claude Code\textsubscript{Haiku 5.5} & \textbf{0.0} & \textbf{0.0} & \textbf{1.0} & 1.1 & \textbf{1.8} & \textbf{0.9} & 0.74 & 0.80 & 80 \\
         & Codex\textsubscript{GPT-5.5} & 1.2 & \textbf{0.0} & 1.7 & \textbf{0.8} & 2.5 & 1.4 & 0.77 & 0.82 & 87 \\
         & RAGLog\textsubscript{Gemini 3.1 Flash-Lite} & 3.1 & \textbf{0.0} & 5.8 & 1.5 & 9.6 & 4.6 & 0.79 & 0.82 & 86 \\
         & RAGLog\textsubscript{GPT-5 Mini} & 6.1 & \textbf{0.0} & 3.6 & 4.6 & 6.6 & 4.1 & 0.80 & \textbf{0.84} & 89 \\
         & RAGLog\textsubscript{Haiku 5.5} & 6.7 & 0.9 & 4.4 & 6.1 & 8.9 & 5.3 & \textbf{0.81} & \textbf{0.84} & \textbf{90} \\
        \bottomrule
    \end{tabular}
    
    \vspace{1mm}
    \footnotesize Conf: Percentage of runs in which the system correctly identifies the confirmed injected fault at the targeted stage (115 in DragonBall, 89 in NQ, 118 in MS MARCO; 322 overall).
\end{table}

\begin{table}[!htbp]
    \centering
    \caption{Run-level detection on injected runs}
    \label{tab:rq2-run}
    \small
    \setlength{\tabcolsep}{3pt}
    \begin{tabular}{@{}lrrrrrrrrrrrrrrrr@{}}
        \toprule
        & \multicolumn{4}{c}{DragonBall} & \multicolumn{4}{c}{NQ} & \multicolumn{4}{c}{MS MARCO} & \multicolumn{4}{c}{All} \\
        \cmidrule(lr){2-5}\cmidrule(lr){6-9}\cmidrule(lr){10-13}\cmidrule(lr){14-17}
        System & P & R & F1 & FPR & P & R & F1 & FPR & P & R & F1 & FPR & P & R & F1 & FPR \\
        \midrule
        Claude Code\textsubscript{Haiku 5.5} & 0.99 & 0.80 & 0.88 & 2.9 & 0.99 & 0.82 & 0.89 & 1.5 & \textbf{1.00} & 0.82 & 0.90 & \textbf{0.0} & 0.99 & 0.81 & 0.89 & 1.4 \\
        Codex\textsubscript{GPT-5.5} & \textbf{1.00} & 0.90 & \textbf{0.95} & \textbf{0.0} & \textbf{1.00} & 0.89 & 0.94 & \textbf{0.0} & 0.99 & 0.86 & 0.92 & 2.1 & \textbf{1.00} & 0.88 & \textbf{0.94} & \textbf{0.7} \\
        RAGLog\textsubscript{Gemini 3.1 Flash-Lite} & 0.96 & 0.92 & 0.94 & 14.3 & 0.91 & 0.95 & 0.93 & 13.8 & 0.94 & 0.89 & 0.92 & 14.9 & 0.94 & 0.92 & 0.93 & 14.3 \\
        RAGLog\textsubscript{GPT-5 Mini} & 0.96 & 0.90 & 0.93 & 11.4 & 0.96 & \textbf{0.96} & \textbf{0.96} & 6.2 & 0.95 & 0.91 & 0.93 & 12.8 & 0.96 & 0.92 & \textbf{0.94} & 9.5 \\
        RAGLog\textsubscript{Haiku 5.5} & 0.95 & \textbf{0.96} & \textbf{0.95} & 17.1 & 0.91 & 0.93 & 0.92 & 12.3 & 0.94 & \textbf{0.96} & \textbf{0.95} & 14.9 & 0.94 & \textbf{0.95} & \textbf{0.94} & 14.3 \\
        \bottomrule
    \end{tabular}

    \vspace{1mm}
    \footnotesize P: precision; R: recall; F1: F1 score; FPR: false-positive rate. \\ Based on 480 injected test runs; a run is faulty if at least one stage is faulty.
\end{table}

Tables~\ref{tab:rq2} to~\ref{tab:rq2-run} show the effectiveness of RAGLog and the two agent baselines on a test set of 480 injected runs. We report stage-level precision, recall, and F1 score in Table~\ref{tab:rq2}, false-positive rates, macro-F1, pooled F1, and detection of confirmed injected faults in Table~\ref{tab:rq2-fpr}, and run-level detection in Table~\ref{tab:rq2-run}. In contrast to natural runs, the agents are competitive on injected runs. \raglogv{Haiku 5.5} and \raglogv{GPT-5 Mini} achieve a pooled F1 score of 0.84, followed by Codex and \raglogv{Gemini 3.1 Flash-Lite} at 0.82 and Claude Code at 0.80. However, these differences are small: only \raglogv{Haiku 5.5} and \raglogv{GPT-5 Mini} significantly outperform Claude Code, both by 0.04, while no RAGLog variant significantly outperforms Codex. For macro-F1, \raglogv{Haiku 5.5} achieves the highest score of 0.81, compared to 0.77 for Codex and 0.74 for Claude Code.

As with natural runs, the main difference is the trade-off between precision and recall. Claude Code and Codex achieve pooled precision scores of 0.95 and 0.94, respectively, with false-positive rates of 0.9\% and 1.4\%. However, their pooled recall is 0.69 and 0.74, as shown in Figure~\ref{fig:pooled-pr}. In contrast, \raglogv{Haiku 5.5} achieves a recall of 0.86, with a precision of 0.81 and a false-positive rate of 5.3\%. Thus, RAGLog identifies more faulty stages but also produces more false positives. The agents perform considerably better on injected runs than on natural runs, possibly because some injected changes, such as an incorrect year in a rewritten query or a truncated embedding query, are more easily identifiable in execution logs. 

\subsubsection{Stage-level effectiveness}
Rewriter and embedder faults are detected effectively by all systems, with F1 scores ranging from 0.86 to 0.97 across the overall results (Table~\ref{tab:rq2}). The differences are more pronounced at later stages. For the generator, \raglogv{Haiku 5.5} achieves an F1 score of 0.69, compared to 0.55 for Claude Code and 0.50 for Codex. The agents detect less than half of generator faults, with recall scores of 0.41 and 0.38, respectively, while RAGLog achieves 0.75. At the retriever, RAGLog achieves F1 scores of 0.68--0.79, compared to 0.75 for Codex and 0.57 for Claude Code. At the reranker, RAGLog achieves F1 scores of 0.75--0.82, compared to 0.76 for Claude Code and 0.73 for Codex; \raglogv{Haiku 5.5} combines a reranker recall of 0.94 with a precision of 0.68. 

\subsubsection{Effectiveness across datasets}
Performance varies across the three datasets (Table~\ref{tab:rq2-fpr}). On DragonBall, Codex and \raglogv{Haiku 5.5} achieve the highest pooled F1 score of 0.83. On NQ, \raglogv{GPT-5 Mini} achieves the highest pooled F1 score of 0.86, compared to 0.84 for Codex and 0.83 for \raglogv{Haiku 5.5}. On MS MARCO, RAGLog performs better, with \raglogv{Haiku 5.5} achieving a pooled F1 score of 0.85, compared to 0.79 for Claude Code and 0.80 for Codex.

\subsubsection{Detection of confirmed injected faults}
Among the 322 test runs in which the judges confirmed a fault at the injected stage, \raglogv{Haiku 5.5} correctly flags that stage in 90\% of runs, compared to 87\% for Codex and 80\% for Claude Code (Table~\ref{tab:rq2-fpr}). The other RAGLog variants achieve detection rates of 89\% for \raglogv{GPT-5 Mini} and 86\% for \raglogv{Gemini 3.1 Flash-Lite}. These results show that RAGLog is particularly effective at identifying the stage where an injected fault occurred, even when its overall pooled F1 is similar to that of the agents.

\subsubsection{Run-level detection}
Of the 480 injected test runs, 333 contained at least one stage-level fault, while 322 had a confirmed fault at the targeted injection stage. At the run level, all systems detect most runs containing faults (Table~\ref{tab:rq2-run}). Codex, \raglogv{Haiku 5.5}, and \raglogv{GPT-5 Mini} achieve the highest F1 score of 0.94, followed by \raglogv{Gemini 3.1 Flash-Lite} at 0.93 and Claude Code at 0.89. \raglogv{Haiku 5.5} achieves the highest recall of 0.95, compared to 0.88 for Codex and 0.81 for Claude Code. However, it also has a substantially higher false-positive rate of 14.3\%, compared to 0.7\% and 1.4\% for the agents. Since every run in this test set underwent fault injection, these false positives occur in runs where the judges found no stage-level fault despite the injection. \raglogv{GPT-5 Mini} offers a better balance in this setting, achieving the same F1 score of 0.94 with a lower false-positive rate of 9.5\%. 

\begin{table}[!htbp]
    \centering
    \caption{Manifestation of the injected faults}
    \label{tab:rq2-propagation}
    \small
    \begin{tabular}{@{}llrrrr@{}}
        \toprule
        Stage & Fault type & Runs & Confirmed & Later stage faulty & Wrong answer \\
        \midrule
        Rewriter  & A wrong year is added                       & 43  & 43 (100\%) & 14 & 7  \\
                  & The entity is replaced by a similar one     & 16  & 16 (100\%) & 10 & 10 \\
                  & A date or qualifier is dropped              & 1   & 1 (100\%)  & 1  & 0  \\
        Embedder  & The query is cut to its first half          & 76  & 76 (100\%) & 18 & 16 \\
                  & The key entity is misspelled                & 74  & 74 (100\%) & 3  & 3  \\
        Retriever & The evidence chunks are not returned        & 60  & 50 (83\%)  & 8  & 49 \\
        Reranker  & The evidence is ranked below the cut-off    & 30  & 20 (67\%)  & 4  & 17 \\
                  & One of several evidence chunks is ranked below the cut-off & 30 & 9 (30\%) & 4 & 9 \\
        Generator & The evidence is placed among similar-entity passages & 150 & 33 (22\%) & -- & 17 \\
        \midrule
        Total     &                                             & 480 & 322 (67\%) & & 128 \\
        \bottomrule
    \end{tabular}
    
    \vspace{1mm}
    \footnotesize Based on 480 test runs: runs per fault type, runs in which the judges confirmed a fault at the injected stage, and, among the confirmed runs, runs in which a later stage was also labeled faulty and runs with a wrong answer.
\end{table}

\subsubsection{Manifestation and propagation of faults}
Table~\ref{tab:rq2-propagation} examines how injected faults manifest and propagate across the 480 test runs. Whether an injection produces a confirmed fault depends strongly on the targeted stage. All rewriter and embedder injections produce faults at their respective stages, compared to 83\% of retriever injections, 48\% of reranker injections, and 22\% of generator injections. Many injections at later stages are absorbed without producing a fault under the judges' criteria. For example, demoting an evidence chunk at the reranker may still leave sufficient evidence in the selected context, while generators may produce correct answers even when relevant evidence is mixed with passages about similar entities. The consequences of confirmed faults also vary by stage. Retriever and reranker faults frequently affect the final answer, with incorrect answers in 49 of 50 and 26 of 29 confirmed cases, respectively. In contrast, many rewriter and embedder faults leave the final answer correct. These cases demonstrate why evaluating only final-answer correctness can miss faults in earlier stages. They also show that faults do not necessarily propagate through the entire pipeline, even when a stage produces an incorrect output.

\begin{answerbox}{Answer to RQ2:}
On injected runs, RAGLog is competitive with the agent baselines, achieving pooled F1 scores of 0.82--0.84 with its three variants, compared to 0.82 for Codex and 0.80 for Claude Code. RAGLog achieves higher recall and correctly identifies the injected stage in up to 90\% of runs with confirmed injected faults, but produces more false positives. It performs best at the generator, the retriever, and the reranker. Among the variants, \raglogv{Haiku 5.5} achieves the highest macro-F1, with GPT-5 Mini offering a lower false-positive rate. The fault-injection analysis further shows that many injections are absorbed and that faults in earlier stages can remain undetected by answer-level evaluation because the final answer is still correct.
\end{answerbox}

\subsection{RQ3: How do the key components of RAGLog contribute to its effectiveness?}\label{sec:rq3}

\begin{table}[!htbp]
    \centering
    \caption{Accuracy (\%) of the API call classification and the log-to-schema transformation}
    \label{tab:rq3-steps}
    \small
    \begin{tabular}{@{}llrrrrrrr@{}}
        \toprule
        & & & \multicolumn{2}{c}{Rules alone} & \multicolumn{2}{c}{LLM alone $^*$} & \multicolumn{2}{c}{Hybrid} \\
        \cmidrule(lr){4-5}\cmidrule(lr){6-7}\cmidrule(l){8-9}
        Step & Set & Calls & Coverage & Correct & Haiku 5.5 & GPT-4.1 mini & Haiku 5.5 & GPT-4.1 mini \\
        \midrule
        Classification & Familiar APIs & 80,778 & 100 & 100 & 100 & 92 & 100 & 100 \\
                       & Unfamiliar APIs & 158 & 96 & 99 & 96 & 91 & 99 & 99 \\
        \midrule
        Transformation & Familiar APIs & 18 & 94 & 100 & 100 & 100 & 100 & 100 \\
                       & Unfamiliar APIs & 124 & 94 & 99 & 98 & 97 & 98 & 98 \\
        \bottomrule
    \end{tabular}
    
    \vspace{1mm}
    \footnotesize $^*$ The LLM-only classification was evaluated on a sample of 51 calls. \\
    Coverage: percentage of calls handled by the rules; Correct: percentage of covered calls that are handled correctly.
\end{table}

\subsubsection{API call classification}
Table~\ref{tab:rq3-steps} shows the effectiveness of the rule-based, LLM-based, and hybrid approaches. On familiar APIs from the five evaluated pipelines, the rules correctly classify all 80,778 calls without requiring the LLM. With the rules disabled, Claude Haiku 5.5 correctly classifies all 51 sampled calls, compared to 47 for GPT-4.1 mini. On 158 unfamiliar API calls, the rules classify 151, of which 149 are correct. The hybrid correctly classifies 156 calls with either model, compared to 151 and 144 for the LLM alone, while sending only seven calls to the LLM. Overall, the hybrid approach provides the best balance of accuracy and efficiency, combining reliable rule-based classification with selective LLM assistance. 

\subsubsection{Log-to-schema transformation}
When the mappings for the five pipelines are rebuilt from scratch, both the hybrid and LLM-only approaches correctly reproduce all 18 stage traces. The rules independently map 17 stages, with only Verba's retriever requiring LLM assistance. On 124 unfamiliar API stage calls, the rules map 116, of which 115 are correct. The hybrid produces correct traces for 121 calls with Claude Haiku 5.5 and 122 with GPT-4.1 mini. The remaining errors involve unusual response formats, such as ranking results identified by IDs rather than positions. Together, these results show that rules handle most classification and transformation tasks, while LLMs resolve cases the rules cannot handle.

\begin{table}[!htbp]
    \centering
    \caption{Ablation results for RAGLog's fault localization components.}
    \label{tab:rq3-ablation}
    \small
    \begin{tabular}{@{}lrrr@{}}
        \toprule
        Configuration & Natural Pooled F1 & Injected Pooled F1 & Natural FPR (\%) \\
        \midrule
        \raglogv{Haiku 5.5} & 0.728 & 0.837 & 4.6 \\
        \midrule
        \multicolumn{4}{@{}l}{\textit{Stage checks}} \\
        \quad Without rule-based checks & 0.712 ($-$0.016) & 0.817 ($-$0.020)$^{\dagger}$ & 4.6 \\
        \quad Without embedding-based checks & 0.723 ($-$0.005) & 0.726 ($-$0.111)$^{\dagger}$ & 4.5 \\
        \quad Without LLM-based checks & 0.453 ($-$0.275)$^{\dagger}$ & 0.667 ($-$0.170)$^{\dagger}$ & 6.6 \\
        \quad Without graded stage check & 0.717 ($-$0.011) & 0.813 ($-$0.024)$^{\dagger}$ & 4.9 \\
        \quad Without graded generator check & 0.725 ($-$0.003) & 0.824 ($-$0.013) & 3.5 \\
        \quad Without generator verification & 0.719 ($-$0.009) & 0.837 (+0.000) & 5.0 \\
        \quad Without evidence flow check & 0.701 ($-$0.027)$^{\dagger}$ & 0.812 ($-$0.025)$^{\dagger}$ & 4.3 \\
        \quad Without statement-level generator check & 0.727 ($-$0.001) & 0.831 ($-$0.006) & 4.1 \\
        \multicolumn{4}{@{}l}{\textit{Ensemble decision}} \\
        \quad Logistic regression only & 0.685 ($-$0.043)$^{\dagger}$ & 0.786 ($-$0.051)$^{\dagger}$ & 4.3 \\
        \quad Gradient-boosted trees only & 0.692 ($-$0.036)$^{\dagger}$ & 0.810 ($-$0.027)$^{\dagger}$ & 4.2 \\
        \quad Extremely randomized trees only & 0.714 ($-$0.014) & 0.816 ($-$0.021) & 3.0 \\
        \quad Equal weights & 0.722 ($-$0.006) & 0.840 (+0.003) & 4.1 \\
        \quad Weights 1/2, 1/4, 1/4 & 0.716 ($-$0.012) & 0.825 ($-$0.012) & 4.7 \\
        \quad Weights 1/5, 3/5, 1/5 & 0.696 ($-$0.032)$^{\dagger}$ & 0.839 (+0.002) & 4.1 \\
        \quad Own-stage measurements only & 0.716 ($-$0.012) & 0.818 ($-$0.019) & 3.2 \\
        \quad Without the false-alarm cap & 0.728 (+0.000) & 0.837 (+0.000) & 4.6 \\
        \bottomrule
    \end{tabular}

    \vspace{1mm}
    \footnotesize $^{\dagger}$ Statistically significant difference from \raglogv{Haiku 5.5}: the 95\% bootstrap confidence interval of the difference excludes zero.
\end{table}

\subsubsection{Fault localization}
Table~\ref{tab:rq3-ablation} shows the contribution of individual fault-localization components, marking differences that are statistically significant. LLM-based checks contribute the most: removing them reduces pooled F1 by 0.275 on natural runs and 0.170 on injected runs. Embedding-based checks are particularly important for injected faults ($-$0.111), while rule-based checks contribute modestly on both test sets ($-$0.016 and $-$0.020), significantly only on injected runs. Among individual checks, removing the evidence flow check reduces pooled F1 by 0.027 on natural and 0.025 on injected runs, highlighting the value of relating evidence across stages; it is the only individual check with a significant effect on both sets. The graded stage check also contributes on both sets, significantly only on injected runs. In contrast, the three generator checks show overlapping contributions: removing any one of them changes pooled F1 by at most 0.013, which is not significant, and removing the statement-level generator check, the most expensive one, reduces pooled F1 by only 0.001 on natural and 0.006 on injected runs.

Replacing the ensemble with logistic regression alone significantly reduces pooled F1 by 0.043 on natural runs and 0.051 on injected runs. Gradient-boosted trees alone also perform significantly worse on both sets ($-$0.036 on natural and $-$0.027 on injected runs), whereas extremely randomized trees alone perform similarly to the ensemble ($-$0.014 and $-$0.021, not significant). Most alternative weightings have little impact, changing pooled F1 by at most 0.012; only counting logistic regression once (weights 1/5, 3/5, 1/5) reduces natural pooled F1 by 0.032. Restricting classifiers to their own-stage measurements reduces pooled F1 by 0.012 on natural and 0.019 on injected runs; both differences point to a benefit of cross-stage measurements, although neither is statistically significant. Removing the false-alarm cap has no effect, as the selected thresholds already satisfy its constraint. Overall, beyond the LLM-based and embedding-based checks, the components provide modest observed benefits rather than evidence that each one is necessary.

\begin{answerbox}{Answer to RQ3:}
The hybrid classification and transformation approaches achieve high accuracy by handling most calls through rules and using LLMs only when those rules cannot resolve a call. For fault localization, LLM-based checks contribute the most, followed by embedding-based checks, particularly for injected faults. The evidence flow check, which relates evidence across stages, provides a smaller but significant improvement, and the ensemble significantly outperforms logistic regression and gradient-boosted trees alone, whereas giving each classifier the measurements of other stages shows a modest benefit that is not statistically significant.
\end{answerbox}

\subsection{RQ4: How efficient is RAGLog at detecting and localizing faults in RAG pipelines?}\label{sec:rq4}

\begin{figure}[!htbp]
    \centering
    \includegraphics[width=\linewidth]{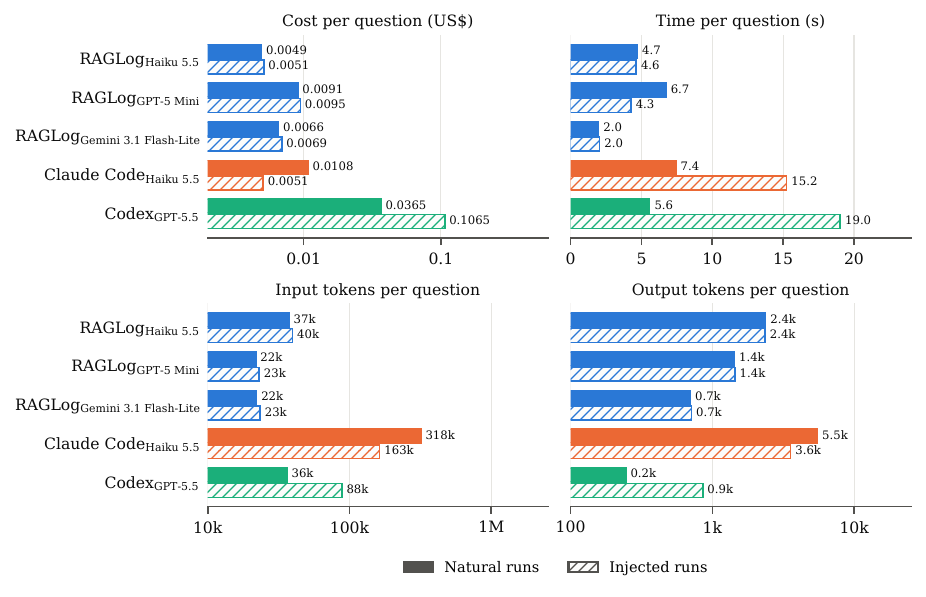}
    \caption{Cost, time, and token usage per question; solid bars show natural runs and shaded bars injected runs}
    \label{fig:rq4-efficiency}
\end{figure}

\begin{figure}[!htbp]
    \centering
    \includegraphics[width=\linewidth]{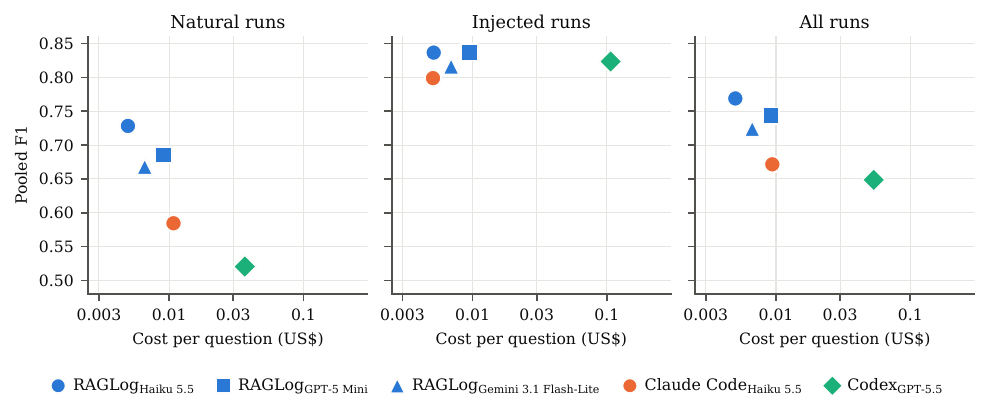}
    \caption{Trade-off between fault-localization effectiveness and cost}
    \label{fig:rq4}
\end{figure}

Figure~\ref{fig:rq4-efficiency} compares the cost, execution time, and token usage per question, while Figure~\ref{fig:rq4} relates cost to pooled F1. Cost and token counts use logarithmic scales. For RAGLog, cost and tokens are averaged per run and time is the median; for the agents, the reported values are session totals divided by the number of questions. 

\subsubsection{Cost and effectiveness}
On natural runs, all three RAGLog variants outperform both agents in pooled F1 while costing less (Figure~\ref{fig:rq4}). \raglogv{Haiku 5.5} achieves the highest F1 (0.73) at US\$0.0049 per question, compared to 0.58 at US\$0.0108 for Claude Code and 0.52 at US\$0.0365 for Codex. On injected runs, the systems achieve similar F1 scores (0.80--0.84), but RAGLog is substantially cheaper than Codex. \raglogv{Haiku 5.5} costs US\$0.0051 per question, matching Claude Code and costing approximately twenty times less than Codex. Across both test sets, \raglogv{Haiku 5.5} achieves the best overall cost-effectiveness, with a pooled F1 of 0.77 at US\$0.0050 per question, compared to 0.67 at US\$0.0094 for Claude Code and 0.65 at US\$0.0535 for Codex. RAGLog's cost is also predictable because each run requires five LLM calls, while the agents' usage depends on their analysis. Classification and transformation add little cost, as rules handle all calls in the evaluated pipelines.

\subsubsection{Tokens}
RAGLog uses 22,000--40,000 input tokens and 700--2,400 output tokens per question. Claude Code uses substantially more input tokens (318,000 on natural runs) because it examines raw logs containing requests, responses, and embedding vectors. However, 87--93\% of the agents' input tokens are cached, reducing their effective cost. Codex uses 36,000--88,000 input tokens per question, with its higher monetary cost also reflecting GPT-5.5's API pricing.

\subsubsection{Time}
\raglogv{Gemini 3.1 Flash-Lite} is the fastest variant, analyzing each run in a median of 2.0 seconds on both test sets, while achieving pooled F1 scores of 0.67 on natural runs and 0.82 on injected runs. This makes it an attractive option when latency is prioritized over maximum effectiveness. \raglogv{Haiku 5.5} takes approximately 4.6--4.7 seconds, offering higher effectiveness at modest additional latency, while \raglogv{GPT-5 Mini} takes 4.3--6.7 seconds. RAGLog's five LLM calls execute in parallel, with local computation taking approximately 0.25 seconds. In contrast, the agents require 5.6--19.0 seconds per question and analyze complete sessions rather than individual runs. RAGLog can therefore provide fault-localization results shortly after each run completes, making it more suitable for monitoring pipelines in production.

\begin{answerbox}{Answer to RQ4:}
RAGLog offers a better cost-effectiveness trade-off than the agent baselines, particularly on natural runs, where all variants achieve higher pooled F1 at lower cost. \raglogv{Haiku 5.5} provides the highest overall effectiveness at approximately US\$0.005 per question, while \raglogv{Gemini 3.1 Flash-Lite} offers the lowest latency (2.0 seconds) with competitive effectiveness. RAGLog's predictable cost and per-run analysis further support its suitability for continuous monitoring.
\end{answerbox}

\section{The RAGLog Tool}\label{sec:tool}
We implemented RAGLog as an open-source tool \footnote{\url{https://github.com/hiraa221/raglog}} for debugging, evaluating, and monitoring RAG pipelines. This section describes how developers connect their pipelines to RAGLog, inspect its findings, and configure the analysis for different use cases.

\subsection{Architecture and Deployment}\label{sec:tool-architecture}
\begin{figure}[!htbp]
    \centering
    \includegraphics[width=0.9\linewidth]{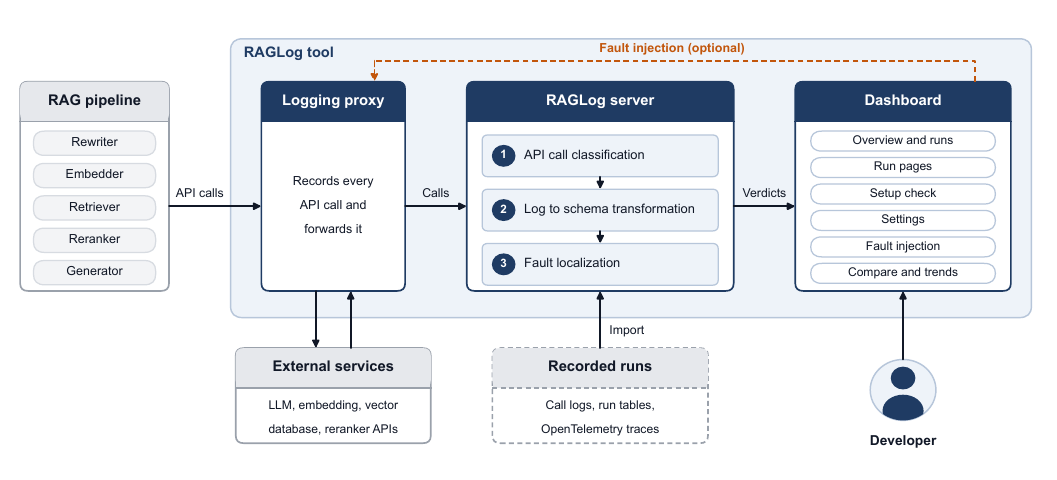}
    \caption{RAGLog Tool Architecture}
    \label{fig:architecture}
\end{figure}

RAGLog consists of three services (Figure~\ref{fig:architecture}): a logging proxy, a server, and a web dashboard. The logging proxy captures outbound API calls from a RAG pipeline (Section~\ref{sec:dataset-logs}). The RAGLog server classifies these calls, transforms them into RAG schema instances, groups them into individual runs, and applies the fault classifiers. The dashboard presents the analysis results and provides controls for configuring and evaluating the pipeline. The services can run together using Docker Compose or as separate processes on a developer's machine. To connect a pipeline, developers route its outbound HTTP traffic through the proxy using standard proxy settings and configure its HTTP clients to trust the proxy's certificate. This requires no changes to the pipeline's source code and works for containerized, local, and server-based deployments. The server analysis runs asynchronously, so it does not delay the pipeline's responses.

RAGLog also supports offline analysis of recorded data. It accepts API call logs from the proxy or a simple call-record format, which undergo the same processing as live traffic. Alternatively, developers can provide tabular runs containing questions, retrieved chunks, generator context, and answers for direct fault localization. RAGLog also accepts OpenTelemetry traces with OpenInference attributes~\cite{opentelemetry,openinference}, mapping their embedding, retrieval, reranking, and LLM spans to the corresponding RAG stages. These options allow RAGLog to analyze previously recorded runs and traces from existing instrumentation without routing traffic through its proxy.

\subsection{Connecting a Pipeline}\label{sec:tool-setup}
When a new pipeline is connected, RAGLog automatically generates the mapping between its API calls and the RAG schema. To help developers verify this mapping, the dashboard provides a setup check that shows which stages were captured, their API endpoints, and the information extracted from the latest run. For example, it displays the embedding input, the number of retrieved chunks, and a preview of the generated answer. It also identifies unclassified calls and provides possible explanations for missing stages, such as stages that are not present in the pipeline. If a stage is captured but its information is not extracted correctly, developers can adjust its mapping directly in the dashboard. RAGLog retains the previous mapping alongside the updated version, allowing developers to inspect their changes.

\subsection{Inspecting Runs}\label{sec:tool-runs}

\begin{figure}[!htbp]
    \centering
    \begin{tabular}{@{}c@{\hspace{0.03\textwidth}}c@{}}
        \includegraphics[height=0.54\textwidth]{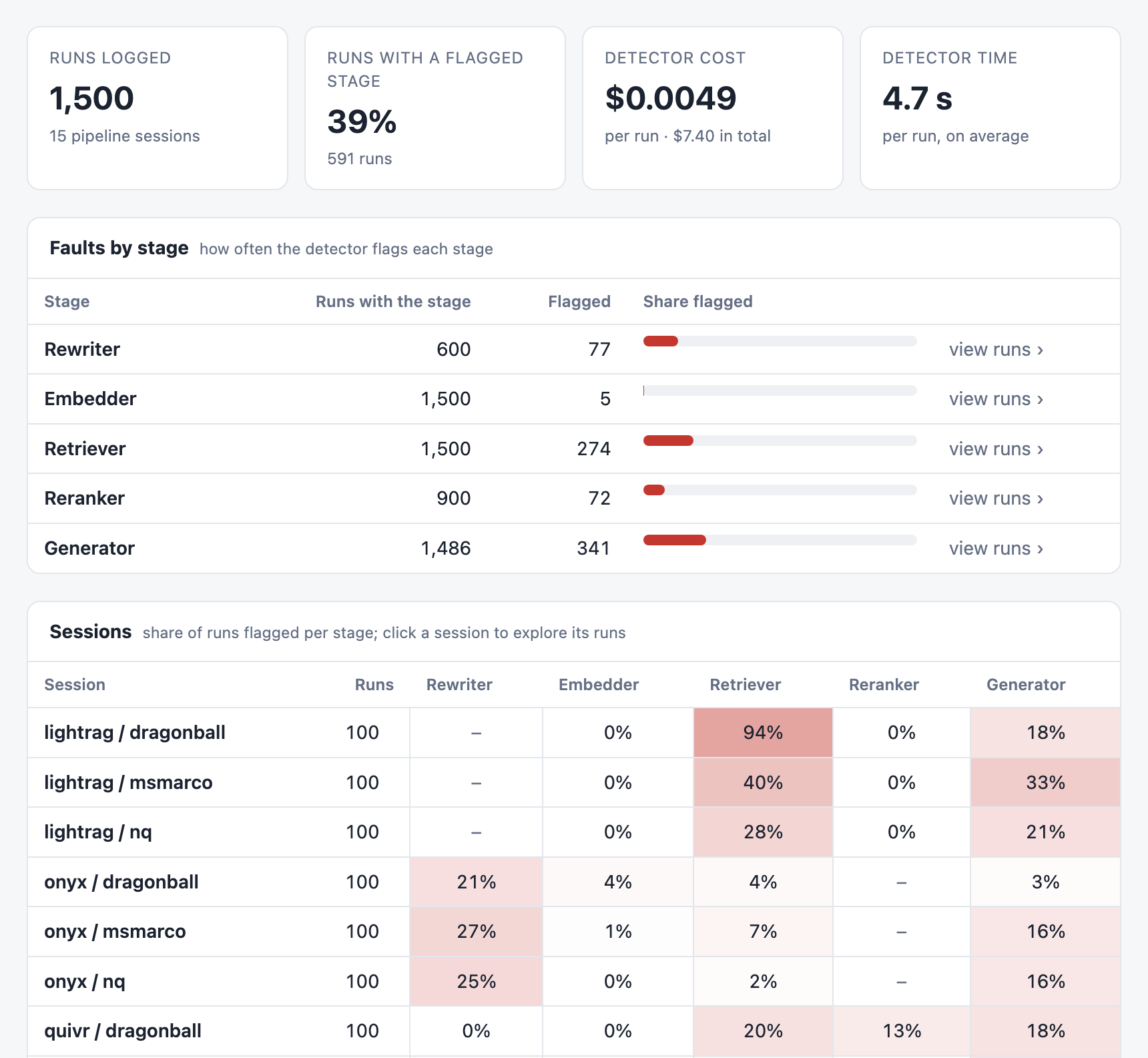} &
        \includegraphics[height=0.54\textwidth]{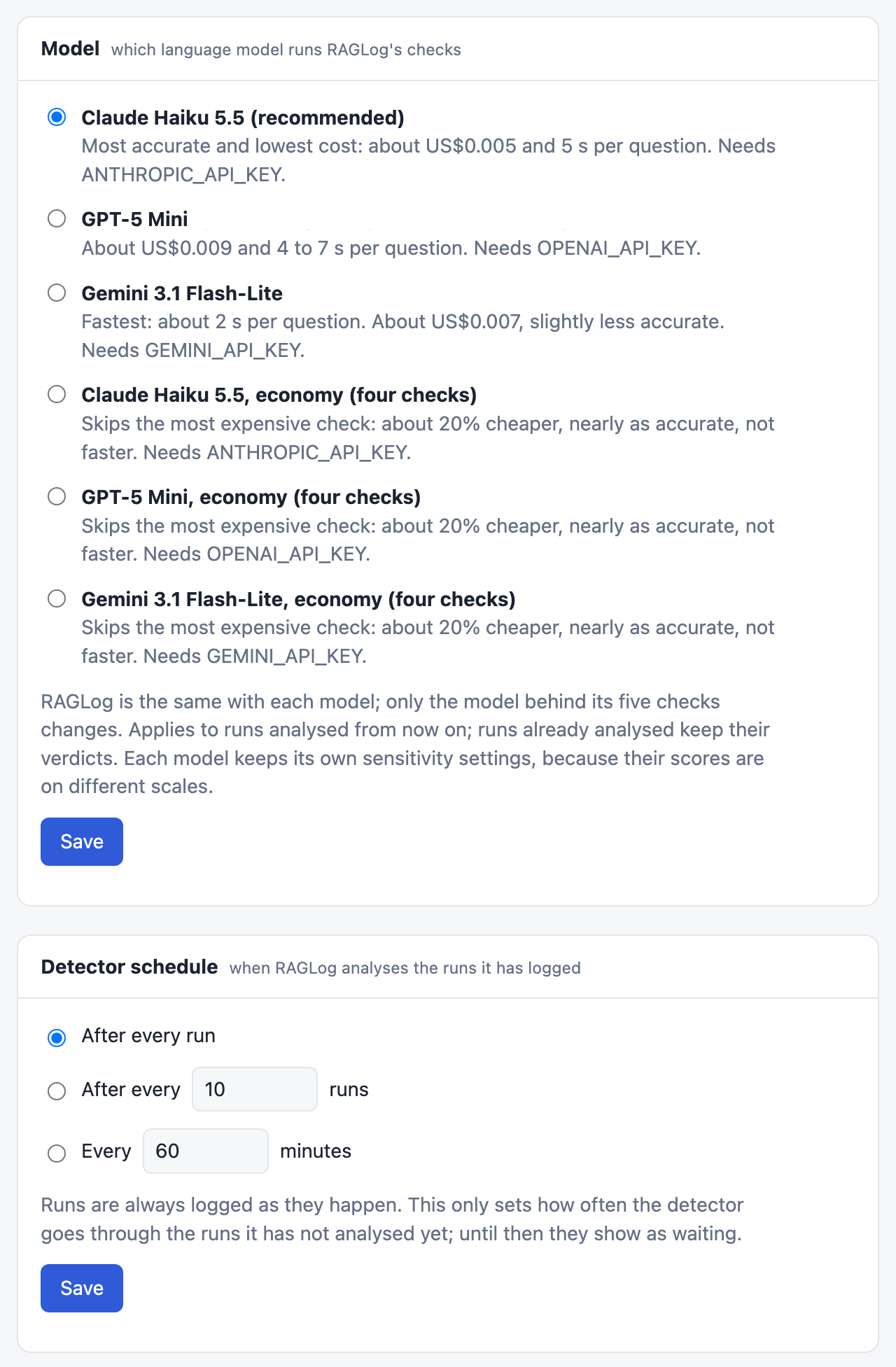} \\[2pt]
        \small (a) Overview & \small (b) Settings
    \end{tabular}
    \caption{Overview of the analyzed runs (a) and the settings for the LLM and the detector schedule (b)}
    \label{fig:tool-overview-settings}
\end{figure}

Once a pipeline is connected, the dashboard provides an overview of its analyzed runs (Figure~\ref{fig:tool-overview-settings}a). It summarizes how often faults are detected, which stages are most frequently flagged, and the time and cost of analysis. Results are organized into sessions, each representing a period or configuration of a pipeline, such as a test execution or software release. Developers can inspect fault rates across sessions and navigate to individual runs. The runs page lists each question alongside its fault verdict, pipeline response time, and RAGLog analysis time.

\begin{figure}[!htbp]
    \centering
    \includegraphics[width=\linewidth]{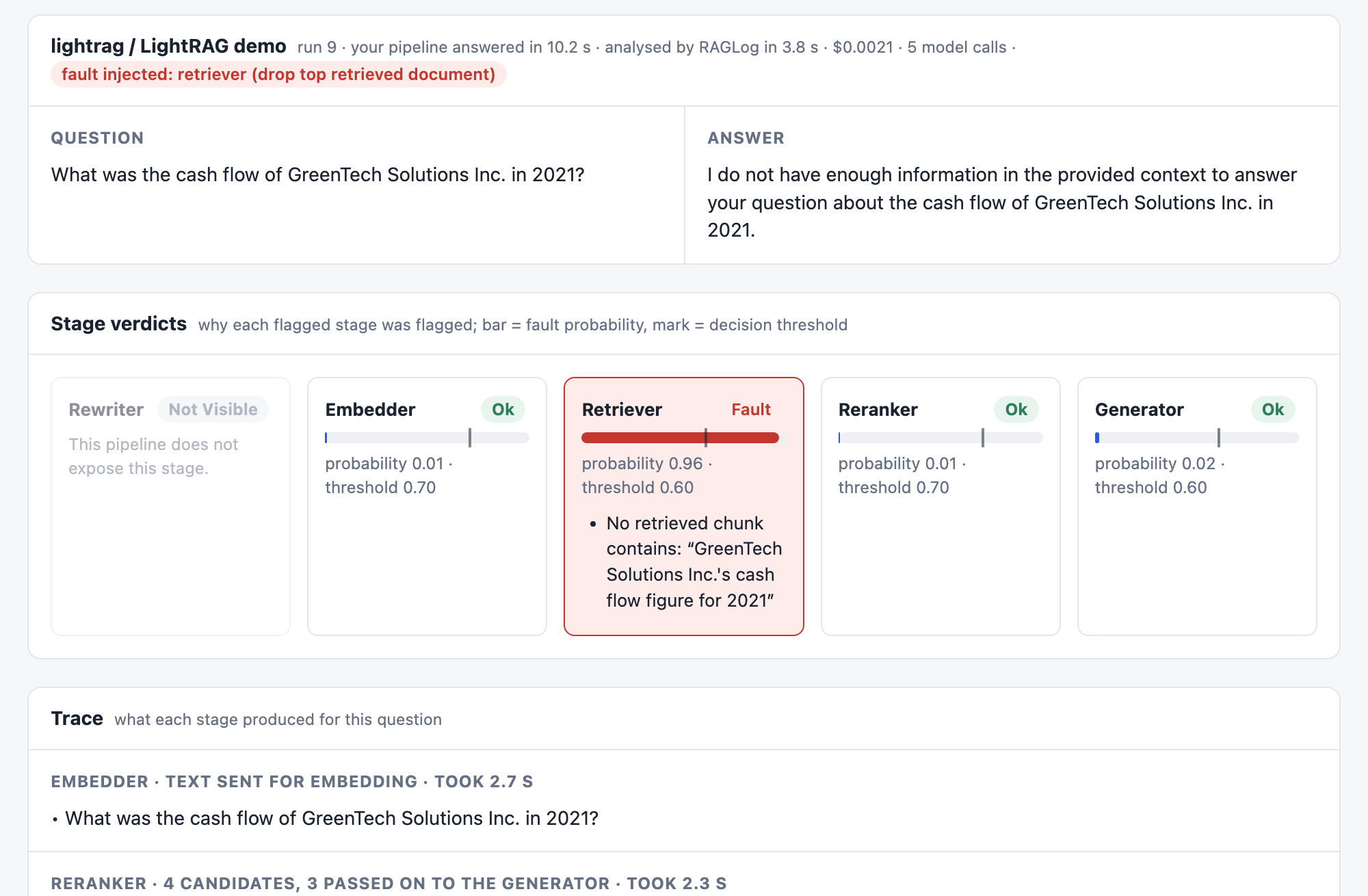}
    \caption{Run page of the RAGLog tool for a LightRAG run with an injected retriever fault: the verdict and reasons
    for each stage, followed by the trace of the run}
    \label{fig:tool-run}
\end{figure}

For each run, RAGLog presents the fault verdict for every stage, its predicted fault probability, and explanation. The dashboard also displays the complete execution trace, including rewritten queries, embedding inputs, retrieved chunks and their scores, generator context, and final answer. These details help developers understand why a stage was flagged and trace the fault to the relevant evidence. For example, Figure~\ref{fig:tool-run} shows a LightRAG run with a retriever fault. The pipeline responds that it lacks sufficient information, while RAGLog identifies the retriever as faulty because none of the retrieved chunks contains the requested cash flow figure. RAGLog can also identify faults that do not affect the final answer.

\subsection{Configuring the Analysis}\label{sec:tool-settings}
Developers can adjust RAGLog's analysis according to their accuracy, latency, and cost requirements (Figure~\ref{fig:tool-overview-settings}b). The main setting is the LLM used for the LLM-based checks. RAGLog supports Claude Haiku 5.5 (the default), GPT-5 Mini, and Gemini 3.1 Flash-Lite, costing approximately US\$0.005, US\$0.009, and US\$0.007 per question, respectively. Each LLM can also be used in an economy mode that omits the statement-level generator check, the most expensive of the five LLM calls. This reduces the cost of an analysis by about 20\%, while keeping pooled F1 within 0.01 of the full configuration in cross-validation. It barely shortens the analysis, because the LLM calls run in parallel and the evidence flow check is usually the slowest. For faster analysis, developers can choose Gemini 3.1 Flash-Lite, which takes about two seconds per run (Section~\ref{sec:rq4}). The released tool uses the configuration evaluated in Section~\ref{sec:results}, but its classifiers are trained on all labeled runs in RAGLog Faults rather than only the development set. For Claude Haiku 5.5, this increases pooled F1 from 0.755 to 0.780 in grouped cross-validation.

RAGLog also allows developers to control when analysis takes place. Runs are always recorded, but fault detection can be performed after every run, after a specified number of runs, or at regular intervals. Immediate analysis supports debugging and monitoring, while batch analysis reduces the load in production. Developers can also set an analysis budget, after which RAGLog stops issuing LLM calls.

To control sensitivity, developers can adjust the fault-classification threshold for each stage. When labeled runs are available, RAGLog suggests thresholds that maximize stage-level F1 or keep the false-positive rate below a specified target, such as 5\%. The dashboard shows the expected precision, recall, and false-positive rate before a threshold is applied. For pipelines retrieving many chunks, developers can also limit analysis to the highest-ranked candidates. Chunks passed to the generator are always retained, but this setting may miss relevant chunks outside the limit, trading some retrieval and reranking fault detection for faster analysis.

\subsection{Testing and Evaluating Pipelines}\label{sec:tool-testing}
Beyond monitoring individual runs, RAGLog supports fault injection and comparative evaluation. The dashboard integrates the live fault injector described in Section~\ref{sec:dataset-injection}, allowing developers to select a stage, a fault type, and the number of questions to affect. Faults are injected into subsequent pipeline runs, so downstream stages respond to the modified outputs. The dashboard identifies injected runs and reports how often RAGLog detects the affected stage. This allows developers to examine how faults propagate through their pipelines and assess the effectiveness of fault localization. Fault types that remove or demote relevant evidence require questions with reference passages and are therefore limited to suitable evaluation datasets.

For evaluation, developers can upload reference answers, enabling RAGLog to assess final-answer correctness alongside stage-level faults at a small additional cost (about US\$0.0001 per run in our tests). Developers can also manually label faulty stages, which RAGLog uses when suggesting detection thresholds. Finally, RAGLog supports comparisons across pipeline sessions. The comparison page displays results from two sessions side by side, such as before and after a pipeline update, and highlights questions whose verdicts have changed.

\section{Discussion}\label{sec:discussion}

\subsection{The Importance of Stage-Level Fault Localization}
Our findings highlight the limitations of evaluating RAG pipelines solely through their final answers. Among the 551 naturally faulty runs in our test set, 202 (37\%) still produced correct answers (Section~\ref{sec:dataset-splits}). Such runs would be considered successful by answer-level evaluation, even though faults occurred in intermediate stages. For example, a retriever may omit relevant evidence or a reranker may discard a useful chunk without affecting the answer to a particular question. These faults can leave the pipeline more vulnerable to failure on other questions, even when their immediate effect is not visible in the output.

RAGLog detects many of these hidden faults. Among the 202 faulty runs with correct answers, \raglogv{Haiku 5.5} correctly identifies at least one faulty stage in 144 runs (71\%), compared with 59 (29\%) for Claude Code and 32 (16\%) for Codex. This is stricter than run-level detection (Table~\ref{tab:rq1-run}), which counts a run as detected whenever any stage is flagged, regardless of whether that stage is actually faulty. For faulty runs that produce incorrect answers, the gap is smaller, with 85\% of runs detected by RAGLog, compared with 63\% and 58\% for Claude Code and Codex, respectively. These findings suggest that the agents may rely more heavily on visible answer failures when identifying faults, whereas RAGLog evaluates intermediate stages regardless of final-answer correctness. Stage-level analysis therefore provides information that answer-level evaluation alone cannot capture \cite{tanhaei2026silent}.

The ablation results further suggest that stage-level fault localization benefits from considering evidence across the pipeline. The evidence flow check, which follows the needed information from retrieval to the answer, is the only individual check whose removal significantly reduces pooled F1 on both test sets, and restricting the classifiers to their own stage’s measurements also reduces pooled F1 on both sets, although not significantly (Section~\ref{sec:rq3}). This is consistent with the interconnected behavior of RAG stages: the output of one stage becomes the input to another, and evidence from downstream stages can help assess whether an earlier stage behaved correctly. These findings support evaluating not only individual stage outputs but also how information flows between stages.

Stage-level verdicts help developers identify where to begin debugging a failed run. In the natural test runs, 71\% of incorrect answers originated from stages before the generator, although answer-level evaluation and hallucination detectors focus on the final output. Among the 349 faulty runs with incorrect answers, the earliest stage flagged by \raglogv{Haiku 5.5} matched the earliest faulty stage in 78\% of runs, compared with 59\% for Claude Code and 55\% for Codex. RAGLog can therefore help developers focus their debugging efforts on where faults originate, such as query rewriting, retrieval, or reranking, rather than making changes to the generator that may not address the underlying problem.

\subsection{Combining Rules, LLMs, and Learned Classifiers}
RAGLog's component evaluations highlight the benefits of assigning different analysis tasks to techniques suited to their characteristics (Section~\ref{sec:rq3}). API call classification and log-to-schema transformation are largely structural tasks, for which rules are often sufficient. The rules correctly classify all calls from the five evaluated pipelines and 149 of 158 calls from unfamiliar APIs. The remaining cases are primarily calls whose role depends on their instructions rather than their endpoint. For example, the same chat API may be used for query rewriting or another generation task. RAGLog uses an LLM to resolve such cases instead of applying it to every call. The hybrid approach correctly classifies 156 of 158 unfamiliar API calls while minimizing reliance on LLMs and their associated costs and variability.

Fault localization presents a different challenge because many faults require semantic judgments about questions, retrieved evidence, and generated answers. Accordingly, removing LLM-based checks produces the largest decline in fault-localization effectiveness of the three evaluated groups of stage checks. However, the LLMs do not make the final fault decisions independently. Their outputs are combined with rule-based and embedding-based measurements through learned classifiers, which also incorporate evidence from other stages. The ablations show smaller, partly significant contributions from these additional measurements and the ensemble compared with the LLM-based checks. Together, the results favor a hybrid design over relying exclusively on deterministic checks or unconstrained LLM judgments.

The choice of LLM for the checks also appears relatively flexible. On natural runs, pooled F1 ranges from 0.67 with Gemini 3.1 Flash-Lite to 0.73 with Claude Haiku 5.5, while the three variants achieve pooled F1 scores of 0.82--0.84 on injected runs (Sections~\ref{sec:rq1} and~\ref{sec:rq2}). Since each LLM-based check performs a narrowly defined assessment and the classifiers are trained on its outputs, these results suggest that RAGLog's effectiveness depends not only on the underlying LLM but also on how its judgments are combined with other measurements. Developers can therefore consider cost, latency, and provider constraints when selecting an LLM, although a different model from the ones provided requires the fault classifiers to be retrained and its effectiveness to be evaluated.

The ablation results also identify opportunities to reduce analysis overhead. Three of the five LLM-based checks assess the generator, and their contributions overlap: removing any one changes pooled F1 by less than 0.02 (Section~\ref{sec:rq3}). Omitting the statement-level generator check reduces cost by about 20\%, with little difference in effectiveness in the evaluated configurations. The released tool offers this configuration as a lower-cost economy mode. We still retain all five checks, since, among all combinations of the three generator checks, this configuration achieves the highest pooled F1 on natural runs, and the combinations differ by at most 0.02 in cross-validation on the development runs. 

\subsection{Specialized Fault Localization and General-Purpose Agents}
The comparison with Claude Code and Codex highlights different trade-offs between specialized fault localization and general-purpose log analysis. On natural runs, the agents have low false-positive rates but miss many faulty stages. Their pooled recalls are 0.46 and 0.40, respectively, compared with 0.78 for \raglogv{Haiku 5.5}, which achieves a pooled F1 of 0.73 against 0.58 and 0.52 for the agents (Section~\ref{sec:rq1}). RAGLog achieves substantially higher recall while maintaining a relatively low pooled false-positive rate of 4.6\%, compared with approximately 1.5\% for the agents. This reflects a trade-off between detecting more faults and generating additional false alarms. Developers can adjust stage-specific thresholds to balance these priorities according to their monitoring requirements. At run level, this trade-off is larger: RAGLog flags about 14\% of fault-free runs, compared with about 5\% for the agents, because a run is flagged when any of its stages is flagged. In monitoring, these false alarms cost a developer the time to inspect a run; stricter thresholds can reduce them at the cost of missing more faults.

The approaches also differ in their analysis overhead. General-purpose agents must interpret the log format, identify individual runs, and extract relevant information before investigating faults. Some of this work is repeated for each session, making the cost sensitive to how logs are divided. For example, Codex costs approximately US\$1.83 per session for natural logs containing 50 questions and US\$1.07 for injected logs containing about 10 questions. Its average cost per question consequently increases from US\$0.037 to US\$0.107 (Section~\ref{sec:rq4}). RAGLog instead uses rules and previously generated mappings to process calls, followed by a fixed set of checks for each run. With Claude Haiku 5.5, its analysis costs approximately US\$0.005 per run and takes around five seconds. These results make RAGLog better suited to repeated analysis, particularly when runs arrive continuously rather than in large batches.

RAGLog also provides structured, repeatable stage-level verdicts, making it suitable for regular monitoring and regression testing. General-purpose agents can undertake more open-ended investigations, including exploring additional log details and generating scripts to examine suspected problems. A practical workflow could therefore use RAGLog to identify potentially faulty stages and reserve agent-based investigation for selected runs requiring deeper diagnosis. This division could reduce repeated analysis overhead while retaining the flexibility of agents for cases that structured checks cannot fully explain.

\subsection{Generalization and Pipeline Adaptation}
Although RAGLog uses a RAG schema to structure logs from different providers into a common representation, its effectiveness decreases on pipelines excluded from training. For \raglogv{Haiku 5.5}, pooled F1 on natural runs falls from 0.73 to 0.67 in the unseen-pipeline evaluation, with similar decreases for the other variants (Section~\ref{sec:rq1}). This reveals an important distinction between processing unfamiliar pipeline logs and accurately identifying faults in those pipelines. RAGLog can classify and transform most unfamiliar API calls, but the learned fault classifiers may still encounter stage behaviors and measurement patterns that are different from those represented in their training data.

Increasing the diversity of labeled runs is one way to address this limitation. Training on additional data already provides some benefit within the evaluated pipelines: when the tool's classifiers are trained using all labeled runs, grouped cross-validation pooled F1 increases from 0.755 to 0.780 for Claude Haiku 5.5 (Section~\ref{sec:tool}). Since RAGLog represents runs using a schema instance, additional pipelines can contribute training examples without requiring changes to the fault-localization architecture. However, the extent to which further training improves performance on unseen pipelines is left for future work. 

These results show that RAGLog transfers to pipelines it was not trained on under our evaluation conditions with open-source pipelines, harmonized settings, benchmark questions, and recorded sessions. They do not establish its effectiveness in production, where queries come from real users, document collections change, and pipelines run continuously; validating RAGLog in a production deployment remains future work.

\subsection{Implications for Fault Injection and Evaluation}
Our fault-injection experiments demonstrate why injecting a disturbance should not automatically be treated as evidence of a fault. Among the 480 labeled injected runs, only 83\% of retriever injections, 48\% of reranker injections, and 22\% of generator injections produced a fault at the targeted stage (Section~\ref{sec:rq2}). Some disturbances were absorbed because alternative evidence remained available or the modified output did not violate the stage's expected behavior. Therefore, treating every injected run as faulty would introduce incorrect ground-truth labels and could penalize a detector for correctly leaving an unaffected stage unflagged.

The distinction is also important when interpreting RAGLog's predictions. RAGLog flags 14\% of injected runs that the judges labeled fault-free, indicating that some checks respond to disturbances even when those disturbances are not confirmed faults. This highlights the difficulty of differentiating unusual stage behavior from behavior that is actually faulty. Fault-injection benchmarks should therefore verify whether the intended fault manifested, rather than equating the injection operation with its outcome. They should also distinguish faults at the injected stage from faults that propagate to downstream stages. Our live fault injector and stage-level labeling procedure support this distinction, allowing fault-localization effectiveness to be evaluated against observed pipeline behavior rather than injection intent.

\subsection{Contributions of the Schema and the Detector}
RAGLog differs from the agent baselines in two ways: it transforms raw logs into structured RAG schema instances and applies targeted stage checks and learned fault classifiers to these instances. To examine the contribution of each, we conducted a pilot experiment in which the agents received RAG schema instances generated by RAGLog instead of execution logs. The task description was adapted to explain the schema fields, while the stage definitions and fault criteria remained unchanged. We selected five test sessions, three containing natural runs (DragonBall on LightRAG, NQ on Onyx, and MS MARCO on RAGFlow) and two containing injected runs (DragonBall on Quivr, MS MARCO on Onyx). These sessions comprised 291 runs, including 150 with faults, and were used to evaluate fault-localization effectiveness (Table~\ref{tab:schema-pilot}).

\begin{table}[!htbp]
    \centering
    \caption{Agent effectiveness and efficiency with raw logs and RAG schema instances in the pilot experiment.}
    \label{tab:schema-pilot}
    \small
    \begin{tabular}{@{}llrrrrrr@{}}
        \toprule
        System & Input Structure & P & R & Pooled F1 & Cost (US\$) & Input tokens & Output tokens\\
        \midrule
        \raglogv{Haiku 5.5} & Raw logs & 0.77 & \textbf{0.86} & \textbf{0.81} & 0.0050 & 39k & 2.3k \\
        Claude Code\textsubscript{Haiku 5.5} & Raw logs & 0.83 & 0.63 & 0.72 & 0.0070 & 171k & 3.6k \\
         & RAG schema & \textbf{0.91} & 0.66 & 0.77 & \textbf{0.0022} & 57k & 1.6k \\
        Codex\textsubscript{GPT-5.5} & Raw logs & \textbf{0.91} & 0.66 & 0.77 & 0.044 & 50k & 0.3k \\
         & RAG schema & 0.78 & 0.71 & 0.75 & 0.026 & \textbf{26k} & \textbf{0.2k} \\
        \bottomrule
    \end{tabular}

    \vspace{1mm}
    \footnotesize P: precision; R: recall.\\ Cost and tokens are per question.
\end{table}
The schema substantially reduces agent resource usage but has a smaller and less consistent effect on fault-localization effectiveness. Claude Code improves mainly on RAGFlow, where in-process retrieval and preprocessed reranker inputs make the raw logs difficult to interpret. By making the reranker's inputs and outputs explicit, the schema improves its reranker F1 from 0.00 to 0.67, while its results on the other sessions only change slightly. Codex, in contrast, achieves higher recall but lower precision with schema instances, resulting in a lower pooled F1.

Despite these improvements, structured inputs do not close the recall gap with RAGLog, particularly for generator faults. RAGLog detects 74\% of generator faults, compared with 15\% for Claude Code and 36\% for Codex when given schema instances. RAGLog systematically applies targeted checks to every run, whereas the agents do not always perform the same detailed analysis, even when given structured inputs. This suggests that RAGLog's effectiveness cannot be attributed to the schema alone; the transformation facilitates log interpretation and reduces analysis cost, while the checks and learned classifiers provide systematic fault localization. However, these findings are preliminary, as the pilot experiment covered only five sessions.
\section{Threats to Validity}\label{sec:threats}

\subsection{Internal Validity}
A primary threat to internal validity is the accuracy of the ground truth. Manually labeling every stage of 3,134 runs was not feasible, so we rely on two LLM judges, which may produce incorrect labels. To mitigate this threat, we use judges from different providers, provide reference answers and evidence, derive stage labels from their factual findings using fixed rules, and employ a third LLM to adjudicate disagreements. Two judges share providers with RAGLog's underlying LLMs (OpenAI and Anthropic), potentially favoring those variants. To reduce this risk, the judges have access to reference evidence, use different prompts and different models, and their labels must agree or undergo adjudication. For the development runs, which are used only for training, we manually reviewed 64 runs with correct answers that the judges labeled as generator faults solely because of unsupported statements. Some involved harmless elaborations, common knowledge, or hedged inferences. Following fixed criteria, we retained a fault when a statement contradicted the retrieved chunks, provided an unsupported answer to the question, or introduced a specific name, number, date, or place. We corrected 19 labels (2\% of development runs). These corrections may affect the learned classifiers, but the test labels remained unchanged. We did not validate the judges’ test labels against independent human annotation, as our manual review covered only development runs. Effectiveness scores are therefore based on these labels, which may contain errors. Overall, the judges agreed on 94\% of stage decisions (Cohen’s $\kappa$ = 0.72), although agreement was lower for the embedder ($\kappa$ = 0.27, with only 10 embedder faults), and the rewriter ($\kappa$ = 0.47).

LLM nondeterminism also affects classification, transformation, fault checks, judges, and agent baselines. RAGLog mitigates its impact through deterministic rules, which correctly handle all API calls in the evaluated pipelines and 149 of 158 unfamiliar calls, and through fault classifiers that combine rule-based, embedding-based, and LLM-based measurements. However, variability in LLM outputs may affect the results. Agent effectiveness may also depend on task instructions. We mitigate this threat by providing them with the same fault definition as the judges, describing the log format, and avoiding prompt tuning on test results.

\subsection{External Validity}
We evaluate five open-source RAG pipelines with different architectures, models, and retrieval backends, using three English-language datasets: DragonBall (finance, law, and medicine), NQ (Wikipedia articles), and MS MARCO (web passages). These pipelines and datasets do not represent all production settings, limiting external validity. Furthermore, the leave-one-pipeline-out evaluation shows reduced effectiveness on unseen pipelines (Section~\ref{sec:rq1}), suggesting that generalization may have limits.

RAGLog covers observable stages after data ingestion. It supports HTTP-based API calls and Weaviate's gRPC calls through a dedicated logger, but not other binary protocols, such as PostgreSQL's protocol~\cite{postgresql}. Although HTTP-based communication is common in RAG pipelines, unsupported protocols may limit applicability. Agentic RAG pipelines, which dynamically select and repeat stages, are also not supported. Effectiveness may further depend on the underlying LLM. We evaluated three LLMs from different providers, but other or newer models may behave differently. Additionally, the local cross-encoder used to select excluded chunks for LLM-based checks was trained on MS MARCO, which is also an evaluation dataset. However, it is only used in pipelines retrieving many chunks (Quivr and RAGFlow) at similar frequencies across all three datasets (37--40\% of runs), reducing the likelihood of a dataset-specific advantage.

\subsection{Construct Validity}
Our definition of a faulty stage considers deviations from the expected output for a given input, even when the final answer remains correct. Alternative definitions could produce different labels. We therefore also report run-level detection, which does not depend on the identified faulty stage. Furthermore, because the judges and RAGLog use the same classified and transformed traces, errors in either step could affect both labels and measurements without being detected. We address this by evaluating API classification and log-to-schema transformation separately in RQ3.

Since injected runs may not fully represent naturally occurring faults, we evaluate natural faults separately and count injected stages as faulty only when confirmed by the judges. We also replaced 36 NQ and MS MARCO questions whose reference answers were independently judged by two LLMs to be unsupported by their source documents.

The agents' measured efficiency depends on session size because they repeat log exploration and script generation for each session. Natural run sessions contain 50 questions, compared to approximately 10 for injected runs, contributing to Codex's cost difference (US\$0.037 versus US\$0.107 per question). We therefore report efficiency separately for both test sets. Larger natural run sessions favor the agents, yet all RAGLog variants remain less expensive. Analyzing individual runs, as required for immediate monitoring, could further increase agent costs. Finally, monetary estimates depend on provider pricing at the time of evaluation and may change in the future.

\subsection{Conclusion Validity}
Some stages have far fewer faults than others, making their individual F1 scores less reliable. For example, natural test runs contain only 10 embedder faults compared to 329 generator faults. As a result, even a few cases from rare stages can have a large impact on F1 scores and macro-F1. We therefore focus on pooled F1, which aggregates decisions across all stages.

To account for variation in the selected test questions, we report bootstrap confidence intervals for comparisons with the agents (Section~\ref{sec:results}). We also assessed variability in LLM outputs by repeating the LLM-based checks of \raglogv{Haiku 5.5} twice on 150 natural test runs (10 per pipeline and dataset). Across the three runs, 95--97\% of stage verdicts were consistent, and pooled F1 varied by only about 0.02, which is much less than RAGLog's advantage over the agents on natural runs. We did not repeat the agent evaluations due to their higher cost, so their run-to-run variability remains unknown. Finally, fault-classifier thresholds were selected on development runs to keep each stage's false-positive rate below 10\%. Although pooled stage-level false-positive rates remain below 10\% for every test dataset, individual stages sometimes exceed this threshold, such as the generator of \raglogv{Haiku 5.5} on natural DragonBall runs (10.8\%). Thus, the development-set constraint does not guarantee the same bound for every stage and dataset.
\section{Related Work}\label{sec:related}

\subsection{Evaluation and debugging of RAG pipelines}
Several techniques exist for improving the quality of RAG pipelines. Raggy \cite{lauro2025rag} is a real-time interactive
visual debugging tool for the indexing, retrieval, and generation steps of RAG pipelines. RAGAS \cite{es2024ragas}, ARES
\cite{saad2024ares}, RAGChecker \cite{ru2024ragchecker}, RAGEval \cite{zhu2025rageval}, and RAGProbe \cite{sivasothy2025ragprobe} are automated
evaluation frameworks for RAG pipelines that generate data or use references to assess the quality of retrieval and
generation with metrics such as correctness, faithfulness, and relevance. FROAV \cite{lin2026froav} is a RAG observability and agent verification framework with
a visual interface for prototyping, integrated evaluation metrics, and correlation between automated and human judgments. Another interactive evaluation system is RAGTrace \cite{cheng2025ragtrace}, which focuses on retrieval and
generation in RAG pipelines through a multi-level analysis, custom metrics, and cross-component interactions to identify potential failures. Observability platforms such as Langfuse \cite{langfuse}, LangSmith \cite{langsmith}, and Phoenix \cite{phoenix} record a trace of every stage and offer evaluators that score individual outputs, but they require instrumentation of the pipeline and leave the analysis of the traces to the developer. Most of these techniques require instrumentation or reference answers and focus on the retrieval and generation stages, and none of them automatically localizes faults to individual pipeline stages. They are well suited for development but may not scale to production environments, where reference answers are unavailable and every run has to be analyzed at a low cost. In contrast, RAGLog locates faults in any stage of a RAG pipeline from its execution logs, without code instrumentation or reference answers, and at a cost that allows its use in production. 

\subsection{Fault detection and localization in RAG pipelines}
The closest to our work are approaches that detect or localize faults in RAG pipelines \cite{leung2026classifying, dong2025rag, jiao2026doctor, sok2025metarag, tanhaei2026silent}. Tanhaei \cite{tanhaei2026silent} proposes a multi-layer quality-gating framework for detecting silent failures across retrieval, context construction, generation, and runtime behavior. The framework demonstrates that answer-level evaluation alone
misses deeper evidence and provenance defects under controlled fault injection. However, it is designed as a pre-release validation gate: its composite score depends on reference evidence sets, reference document identifiers, answerability labels, and citation mappings that are unavailable in live deployments, in addition to instrumented
runtime metadata not exposed by standard RAG APIs. RAGLog instead detects and localizes faults without reference data, from API call logs. Leung et al.~\cite{leung2026classifying} introduce RAGEC, which defines 16 fault types across the chunking,
retrieval, reranking, and generation stages and automatically classifies the stage and fault type responsible for an incorrect response. RAGEC first compares the generated response against a reference answer and documents to identify incorrect responses. For each identified fault, it then uses reference chunks and intermediate pipeline outputs to
determine the earliest stage that exhibits erroneous behavior. RAGEC therefore supports the post-hoc classification of known faults when reference data are available. In contrast, RAGLog determines whether a fault has occurred and localizes the faulty
stage without reference answers or reference evidence at run time, and without knowing that an answer is incorrect.

RAG-Critic \cite{dong2025rag} trains a critic model, based on a hierarchical taxonomy of RAG faults, that evaluates a response and attributes its faults to retrieval or generation, and uses this feedback to correct the response. It operates without reference answers, but distinguishes only retrieval from generation faults, without separating the steps before generation, and only reports faults where it judges the final answer to be incorrect. Doctor-RAG
\cite{jiao2026doctor} diagnoses agentic RAG runs that are known to have failed, identifies the earliest failing step of the trajectory, and repairs the run from that step. AgenticRAG-FP \cite{pothuru-2026-failures} is a benchmark for causal failure attribution in agentic RAG, which injects a fault at a known retrieval hop and evaluates whether diagnostic methods identify that hop; it attributes failures to hops of a multi-hop trajectory rather than to the stages of a pipeline. Sok et al.~\cite{sok2025metarag} propose MetaRAG, a reference-free, black-box approach for detecting hallucinations in RAG-generated answers. MetaRAG decomposes an answer into atomic factoids, generates synonym and antonym mutations, and verifies these mutations against the retrieved context. Although
MetaRAG shares RAGLog's reference-free setting, it focuses exclusively on the grounding of the generator's answer, and it cannot determine whether an earlier rewriter, embedder, retriever, or reranker produced faulty context. RAGLog instead
analyzes the information flow across five pipeline stages to detect and localize faults in all of them, including faults that propagate to later stages and faults that leave the final answer correct.

\subsection{Fault localization in LLM and agentic systems}
Fault localization techniques also exist for general LLM and agentic systems. Notable ones among these are RAFFLES \cite{zhu2026raffles} and CHIEF \cite{wang2026flat}. RAFFLES locates faults in LLM systems with multiple components by using an LLM judge to evaluate all components and custom evaluators to assess the quality of the faults it detects and the rationale provided by the judge. CHIEF organizes a complex execution trace into a structured causal map and then traces backward through it to locate the root cause of a failure. AgentSight \cite{zheng2025agentsight} is an
instrumentation-free observability framework for agentic systems. It intercepts TLS-encrypted LLM traffic to extract
semantic intent, monitors system behavior through kernel events, and links these two streams across processes to detect
prompt injections, resource-wasting reasoning loops, and coordination bottlenecks in multi-agent systems. Chan et al.
\cite{chan2024visibility} discuss categories and potential implementations of visibility into AI agents, including agent
identifiers, real-time monitoring, and activity logging. More generally, log analysis is a well-established technique for debugging and fault localization in software systems \cite{chen2019empirical, he2018identifying}. These techniques, however, do not understand the roles of the stages of a RAG pipeline; they cannot check whether the information that a question needs was retrieved, passed on by the reranker, and used by the generator, which is where most RAG faults become visible. 

\subsection{Fault injection and datasets}
Fault injection is an established technique for evaluating the robustness of software and deep-learning systems \cite{hsueh1997fault, ma2018deepmutation}. For RAG pipelines, RAGProbe \cite{sivasothy2025ragprobe} generates question scenarios to expose failures, while Tanhaei \cite{tanhaei2026silent} injects controlled faults to demonstrate that answer-level evaluation can miss faults in earlier stages. In contrast, our fault injector modifies the input or output of a single stage during pipeline execution by intercepting API calls, without changing the pipeline's code. This allows subsequent stages to process the modified data, enabling us to examine how faults propagate or are absorbed. Existing RAG datasets primarily label answer correctness or hallucinations \cite{niu2024ragtruth,ru2024ragchecker}, rather than identifying the faulty pipeline stages. RAGLog Faults addresses this gap by providing stage-level fault labels for both natural and injected runs across five real RAG pipelines.

\section{Conclusion and Future Work}\label{sec:conclusion}
We presented RAGLog, a reference-free approach for detecting and localizing faults across RAG pipeline stages using execution logs, without code instrumentation. RAGLog combines rules and LLMs to classify API calls and transform logs into a RAG schema instance. It then uses rule-based, embedding-based, and LLM-based checks with an ensemble of learned classifiers to identify faulty stages, allowing pipelines from different providers to be analyzed without manual configuration.

To evaluate RAGLog, we developed a logging proxy and a live fault injector. We also collected RAGLog Faults, a dataset of 3,134 runs from five open-source RAG pipelines (LightRAG, Onyx, Quivr, RAGFlow, and Verba) across three public datasets (DragonBall, Natural Questions, and MS MARCO), with stage-level fault labels. We compared RAGLog against two general-purpose coding agents, Claude Code and Codex, which analyzed the same execution logs. With Claude Haiku 5.5 as its underlying LLM, RAGLog achieved a pooled F1 of 0.73 on natural runs, outperforming Claude Code (0.58) and Codex (0.52). On injected runs, all three achieved comparable effectiveness (0.84, 0.80, and 0.82, respectively). RAGLog also achieved overall lower cost and latency than both agents, even when compared with Claude Code using the same underlying LLM. RAGLog with Claude Haiku 5.5 analyzed each run in approximately five seconds at US\$0.005.

Beyond these comparisons, our results provide insight into RAGLog's effectiveness and the importance of stage-level fault localization. More than one-third of naturally faulty runs still produced correct answers, showing why evaluating only the final answer is not sufficient. RAGLog addresses this by combining measurements across stages. The ablation study confirms the importance of LLM-based checks and suggests a smaller contribution from cross-stage information. Its fault-localization effectiveness also remained relatively consistent across three variants using LLMs from different providers. Meanwhile, the hybrid classification and transformation approach handled most API calls through rules, limiting LLM assistance to unfamiliar cases. We release RAGLog as an open-source tool with a dashboard for debugging, evaluation, and monitoring, together with the RAGLog Faults dataset.

In the future, we plan to expand RAGLog's training data across additional pipelines, providers, and domains. We plan to extend the fault injector to support more diverse fault types, providing broader training and evaluation data. We also plan to include human annotations for a sample of runs to assess agreement with the LLM labels. In addition, we plan to consolidate overlapping generator checks to reduce analysis costs and extend RAGLog to data ingestion stages and agentic RAG pipelines.

\section*{Acknowledgments}
Naveed is supported by a Faculty of IT Post-graduate scholarship. Grundy and Haggag are supported by ARC Laureate Fellowship FL190100035. Haggag is also supported by a National Intelligence Post-doctoral Fellowship. This work is also partly supported by ARC Discovery Project DP200100020. ChatGPT and Claude were used to assist with plotting scripts and academic writing. Claude was also used to assist with portions of the RAGLog implementation and evaluation scripts. The authors reviewed the AI-assisted outputs and take full responsibility for the content of this work.

\section*{Data Availability}
The RAGLog Faults dataset is available at \url{https://doi.org/10.57967/hf/11214}. \noindent The RAGLog Tool is available at \url{https://github.com/hiraa221/raglog}. A complete replication package is available at \url{https://doi.org/10.5281/zenodo.23792939}.


\balance 
\bibliographystyle{ACM-Reference-Format}
\bibliography{sample-base}

\end{document}